\documentclass{article}

\usepackage{arxiv}
\usepackage{scalerel}
\usepackage[utf8]{inputenc} 
\usepackage[T1]{fontenc}    
\usepackage{hyperref}       
\usepackage{url}            
\usepackage{booktabs}       
\usepackage{amsfonts}       
\usepackage{nicefrac}       
\usepackage{microtype}      
\usepackage{lipsum}
\usepackage{mathtools,halloweenmath}		
\usepackage{graphicx}
\usepackage{amssymb}
\usepackage{dsfont}
\usepackage{doi}
\usepackage{amsthm}
\usepackage{lineno}
\theoremstyle{plain}
\newtheorem{theorem}{Theorem}

\newtheorem{corollary}{Corollary}
\newtheorem{proposition}[theorem]{Proposition}

\theoremstyle{definition}

\theoremstyle{remark}
\newtheorem{remark}{Remark}

\title{Kuramoto Oscillators and Swarms on Manifolds for Geometry Informed Machine Learning}

\author{Vladimir Ja\' cimovi\' c \\
	Faculty of Natural Sciences and Mathematics\\
	University of Montenegro\\ 
	Cetinjski put bb., 81000 Podgorica\\ 
	Montenegro\\
	\texttt{vladimirj@ucg.ac.me} \\
}

\renewcommand{\shorttitle}{Learning non-Euclidean data via Kuramoto models}

\hypersetup{
pdftitle={A proposal on learning non-Euclidean data via Kuramoto model and its extensions},
pdfsubject={-},
pdfauthor={Vladimir Ja\' cimovi\' c},
pdfkeywords={Kullback-Leibler divergence, Fisher information metric, evolutionary games, information-geometric optimization, Fisher speed},
}

\begin{document}
\maketitle

\begin{abstract}

We propose the idea of using Kuramoto models (including their higher-dimensional generalizations) for machine learning over non-Euclidean data sets. These models are systems of matrix ODE's describing collective motions (swarming dynamics) of abstract particles (generalized oscillators) on spheres, homogeneous spaces and Lie groups. Such models have been extensively studied from the beginning of XXI century both in statistical physics and control theory. They provide a suitable framework for encoding maps between various manifolds and are capable of learning over spherical and hyperbolic geometries. In addition, they can learn coupled actions of transformation groups (such as special orthogonal, unitary and Lorentz groups). Furthermore, we overview families of probability distributions that provide appropriate statistical models for probabilistic modeling and inference in Geometric Deep Learning. We argue in favor of using statistical models which arise in different Kuramoto models in the continuum limit of particles. The most convenient families of probability distributions are those which are invariant with respect to actions of certain symmetry groups.

\end{abstract}

\keywords{Geometric Deep Learning, geometric Riccati equations, directional statistics, non-Euclidean data}

\section{Introduction}\label{sec:1}

Machine Learning (ML) is, to a great extent, a science of inferring models and patterns from data. From that point of view, its core objective consists in learning optimal (according to a certain criterion) mappings between spaces. For several decades these mappings have been dominantly encoded using artificial neural networks with different topologies \cite{GBC}. The spaces have almost always been assumed Euclidean or equipped with some flat metric. The data have been represented by points in Euclidean spaces or in finite sets.

An enormous progress in ML and Data Science in XXI century led to the growing understanding that a great deal (possibly, majority) of data sets have inherent non-Euclidean geometries. This fact has been mostly neglected in ML until very recently. Only the last decade brought systematic research efforts focused on geometric-sensitive architectures of neural networks (NN's). 

In parallel, traditional ways of designing artificial NN's are being reexamined and enriched by new ideas. Diversity of applications and conceptual complexity of ML problems motivated investigations of new architectures. Over the centuries mathematicians elaborated various ways of encoding maps between Euclidean spaces or Riemannian manifolds. The corresponding theories have been established before the advent of ML, and now provide a solid theoretical background for its future developments. 
Following an explosive expansion of ML applications and practices, there is a huge backlog of theoretical work to be done. Mathematical foundations of ML are being actively reconsidered and expanded. Certain fields of mathematics that have been almost invisible in ML until very recently are now actively exploited with a great potential for future applications. The examples include Riemannian Geometry, Game Theory and Lie Group Theory - to name just a few.

Systematic approaches in ML must be based on well established theories and well understood models. The choice of adequate models and appropriate data representations appears to be the key issue. An appropriate choice greatly reduces the dimension (number of parameters), increases the efficiency of algorithms and, equally important, improves their transparency. 

The main goal of the present paper is to point out a broad class of models which constitute a powerful theoretical framework for encoding geometric data. These models describing collective motions of interacting particles have been studied in Science for almost half of a century from various points of view. In physics of complex systems they are known as {\it Kuramoto models} \cite{Kuramoto} (including generalizations to higher-dimensional manifolds \cite{MTG,HKK}) and {\it Viscek models} \cite{VCB-JCS}. In systems theory they are said to be {\it (anti-)consensus algorithms} on manifolds \cite{SS,Sarlette}. Finally, in Engineering they are sometimes referred to as {\it swarms} on manifolds \cite{Olfati-Saber,MPG}. All these models fit into the unifying mathematical framework that we refer to as {\it systems of geometric Riccati ODE's}, as will be explained in sections 3 and 4.

Our exposition will be focused on the following questions:

1. Which kinds of mappings can be encoded by collective motions of Kuramoto oscillators/swarms on manifolds?

2. Which symmetries/patterns can be learned using these dynamics? 

3. Which statistical models are associated with these dynamics and how can they be used in statistical ML over manifolds?

4. Which problems can be efficiently solved using such models?

5. How these models can be trained?

Our proposal on using swarms/Kuramoto oscillators in ML is inspired by some recent developments in theoretical ML which will be mentioned in Section 2. Section 3 is devoted to classical Kuramoto models (i.e. models describing collective motions of the classical phase oscillators) and their potential applications to learning coupled actions of transformation groups, as well as data on circles, tori and hyperbolic multi-discs. Section 4 contains an overview of (generalized) Kuramoto models that describe collective motions on spheres, Lie groups and other manifolds. In Section 5 we present families of probability measures over Riemannian manifolds which provide appropriate statistical models for probabilistic ML algorithms over non-Euclidean data sets. Some of these families are generated by the corresponding swarming dynamics. Connections with directional statistics will be particularly emphasized. In Section 6 we clarify how swarms can be used for supervised, unsupervised and reinforcement learning over Riemannian manifolds. In Section 7 we analyze some illustrative geometric ML problems in low dimensions, thus supporting our main points. Finally, Section 8 contains some concluding remarks and an outlook for the future research efforts.

\section{Some recent trends in theoretical ML}

Our proposal on ML via swarms on manifolds builds upon the combination of four research directions which recently had a great impact on the field. {\it Control-theoretic ML} investigates new architectures of neural networks (NN's) for Deep Learning (DL) by encoding maps into continuous-time dynamical systems, based on mathematical theories of ODE's and optimal control. {\it ML through probabilistic modeling and inference} aims to encode uncertainties using probability measures. {\it Geometric ML} explores intrinsic geometric features of the data, embeds the instances into Riemannian manifolds and infers the curvature and symmetries hidden in data sets. Finally, {\it Physics Informed ML} leverages laws of Physics (such as conservation laws, time-space symmetries, MaxEnt principle) to design efficient and transparent ML algorithms.

The literature on each of these directions is vast and constantly growing. We do not even try to provide a comprehensive or representative (in any sense) list of references.   

\subsection{Deep Learning via continuous-time controlled dynamical system}

In 2017. Weinan E proposed new architectures of NN's realized through the continuous-time controlled dynamical systems \cite{E}. This proposal was motivated by the previous observations that NN's (most notably, ResNets) can be regarded as Euler discretizations of controlled ODE's. In parallel, a number of studies \cite{LCTE,LT,Recht} enhanced and expanded theoretical foundations of ML by adapting classical control-theoretic techniques to the new promising field of applications.

This line of research resulted in a tangible outcome which was named Neural ODE \cite{CRBD}. The underlying idea is to formalize some ML tasks as optimal control problems. In fact, deep limits of ResNets with constant weights yield continuous-time dynamical systems \cite{AN}. In such a setup weights of the NN are replaced by control functions. Training of the model is realized through minimization of the total error (or total loss) using the Pontryagin's maximum principle. Backpropagation corresponds to the adjoint ODE which is solved backwards in time.   

A similar way of encoding maps underlies the concept of continuous-time normalizing flows \cite{PNRML}. Normalizing flows are dynamical systems, usually described by ODE's or PDE's. These systems are trained with the goal of learning a sequence (or a flow) of invertible maps between the observed data originating from an unknown complicated target probability distribution and some simple (typically Gaussian) distribution. Once the normalizing flow is trained, the target distribution is approximated. The model is capable of generalizing the observed data and making predictions by sampling from the simple distribution and mapping the samples along the learned flow.

We have mentioned two concepts (neural ODE and normalizing flows) that recently had a significant impact. Their success reflects the general trend of growing interest in control-theoretic point of view on ML. Most of theoretical advances in Reinforcement Learning (RL) rely on Control Theory (CT) \cite{LT,Recht}. As theoretical foundations of RL are being established, the boundary between RL and CT is getting blurred. 

\subsection{Probabilistic modeling and inference in DL}

Learning in general can be viewed as a process of updating certain beliefs about the state of the world based on the new information. This abstract point of view underlies the broad field of {\it Probabilistic ML} \cite{Ghahramani}. In this Subsection we mention certain aspects of this field which are the most relevant for the present study.

The general idea of updating beliefs can be formalized as learning an optimal (according to a certain criterion) probability distribution. This further implies that implementation of probabilistic ML algorithms involves optimization over spaces of probability distributions. Therefore, gradient flows on spaces of probability measures \cite{CHHRS,AGS} are essential ingredients of probabilistic modeling in ML. The notion of gradient flow requires the metric structure. The distance between two probability measures should represent the degree of difficulty to distinguish between them provided that a limited number of samples is available. Metric on spaces of probability measures are induced by the Hessians of various divergence functions \cite{AJLS,Barndorff-Nielsen}. The classical (and parametric invariant) choice is the {\it Kullback-Leibler divergence} (KL divergence), also referred to as {\it relative entropy}. This divergence induces the Fisher (or Fisher-Rao) information metric on spaces of probability measures thus turning them into statistical manifolds \cite{AJLS}. When optimizing over a family of probability distributions, Euclidean (or, so called, "vanilla") gradient is inappropriate. Ignoring of this fact, leads to inaccurate or incorrect algorithms. Instead, one should use the gradient w.r. to Fisher information metric, which is named {\it natural gradient} \cite{Amari,MP1,Martens}. In RL this must be taken into account when designing stochastic policies. In particular, well known actor-critic algorithm has been modified in order to respect geometry of the space of probability distributions \cite{PS}.

Another way of introducing metric on spaces of probability distributions is the Wasserstein metric. Fokker-Planck equations are gradient flows in the Wasserstein metric. The potential function for these flows is the KL divergence between the instant and (unknown) stationary distribution \cite{JKO}. Yet another metric sometimes used in ML is the Stein metric \cite{LiuWang}.

\subsection{Deep Learning in non-Euclidean spaces}

A great deal (probably, majority) of data sets are naturally represented in non-Euclidean geometries. This fact has been widely recognized in ML only recently, motivating research efforts in non-Euclidean data representations and ML algorithms over curved spaces. 
Inferring curvature and symmetries encoded in data sets is of the crucial importance in many tasks.

The necessity of geometric methods in ML is easy to justify. It is apparent even in basic setups. In order to support this point, we provide three illustrative cases.

A) When learning rotations in the $3$-dimensional space, traditional NN architectures, which involve Euclidean addition and averaging of vectors, are inappropriate.  

B) When optimizing over a certain family of probability distributions (say, for inference problems), it is advisable to take into account the intrinsic geometry of this specific family. For instance, Gaussian policy parametrization is typically used in stochastic RL algorithms over continuous strategy sets. However, intrinsic geometry of the family of Gaussian distributions ${\cal N}(a,\Sigma)$ is hyperbolic. If one applies standard Euclidean gradient descent to the problem of learning parameters $a$ and $\Sigma$, it is likely that the algorithm will perform incorrectly and learn the matrix $\Sigma$ which is not positive definite. This does not correspond to any probability distribution. Therefore, one should adapt the gradient descent by taking into account geometry of the manifold ${\cal N}(a,\Sigma)$.       

C) Although the Gaussian family usually provides the most convenient statistical model for probabilistic ML algorithms over Euclidean spaces, it is not suitable when learning orientations in space. Data of this kind one require families of probability distributions over the sphere.

Inferring the curvature and symmetries hidden in data sets is a central problem of geometric ML. Roughly, these sets can be classified into spherical data (data embedded into spaces with strictly positive curvature), Euclidean data (data with zero curvature) and hyperbolic data (data with strictly negative curvature). Of course, such a classification is highly oversimplifying as the majority of real-life big data have mixed curvature \cite{GSGR,WHPD}

An apparent example of spherical data are orientations in the Euclidean space. Another less obvious example is the space of categorical distributions (probability distributions over a finite set) equipped with Fisher information metric. The natural gradient update (i.e. update w. r. to Fisher metric) on the manifold of categorical distributions amounts to optimization in spherical geometry.

On the other hand, hyperbolic data are even more ubiquitous in Science. A great deal of data sets have some (possibly hidden) hierarchical structure. Such data are naturally embedded into manifolds with hyperbolic geometry. For example, the power law (also known as the Pareto-Zipf law) for degrees of nodes in complex networks implies hyperbolic geometry, and vice versa \cite{KPKVB,MTCBC}. Other examples of inherently hyperbolic data are common in word embeddings and natural language processing \cite{TBG,LW}, molecular structures \cite{Poleksic}, Gaussian distributions \cite{CSS}. In general, one might claim that most of biological data have inherently hyperbolic geometry.

Optimization on manifolds is a young subdiscipline within the broad field of mathematical optimization. Although particular problems of this kind (such as Wahba's problem) occasionally appeared in the literature for a long time, systematic approaches have been elaborated in XXI century \cite{AMS,Boumal}. Nowadays, advances in the theory of optimization on manifolds are, to a great extent, motivated by applications to geometric ML. 
For some examples, we refer to ML algorithms based on optimization over hyperbolic \cite{GBH,NK,LLSZ} or spherical \cite{FHPS,CGKW} geometries. Recently, novel architectures of spherical \cite{CGKW} and hyperbolic \cite{GBH} NN's have been proposed for dealing with geometric data.

In parallel, probabilistic modeling in geometric ML exploits statistical models over Riemannian manifolds. This motivated a growing interest in applications of directional statistics to ML\cite{Sra,BDGS,NYFK}. Directional statistics is the subdiscipline within general statistics and probability, which deals with observations on compact Riemannian manifolds. Classical and probably the most comprehensive reference on this field is the book of Mardia and Jupp \cite{MJ}. 

Another approach to probabilistic modeling in geometric ML is provided by normalizing flows over Riemannian manifolds. Some researchers\cite{MN,RPRAKSC,GRM,BSLPH} reported experiments with normalizing flows over spheres, tori and other manifolds for the density estimation problem.

Summarizing, encoding geometric features of data in deep learning models evolved into the emerging field of Geometric Deep Learning \cite{BBCV}.

\subsubsection{Learning (coupled) actions of transformation groups}

In Geometric DL there is an important class of problems, where the goal is to learn transformations, such as rotations in the $d$-dimensional space, conformal mappings, groups of isometries, etc. Such problems arise in robotics (movement prediction \cite{UTP} and imitation learning \cite{OPNBAP}), in analysis of facial expressions \cite{ZLCMG}, computer vision \cite{Lui}, etc. The corresponding algorithms rely on optimization over Lie groups \cite{LZZ,CMBP}, as well as on learning probability distributions over Lie groups \cite{FHDF,JA,CW}.

One conceptual approach to problems of this kind is the longstanding idea of NN's with non-Euclidean neurons (and possibly weights) \cite{ZWM,BS}. More recently, this line of reasoning resulted in novel architectures, named equivariant neural networks \cite{CW2}. Equivariant NN's are designed in such a way to ensure that outputs are transformed consistently under symmetry transformations of the inputs. 

Very recently, several researchers reported on pioneering efforts in RL with non-Euclidean spaces of states (and actions). So far, robotics is the dominant field of applications. There are two possible approaches to stochastic policies in RL problems of this kind. Some researchers proposed policy parametrizations on the tangent space using standard statistical models (typically the Gaussian family), and projecting the learned policies onto manifolds via exponential map \cite{ASSAKA-D}. The second approach consists in parametrization of policies using families of distributions over Riemannian manifolds, thus employing results from directional statistic \cite{JA}. 


\subsection{Physics Informed ML}

The term {\it physics informed ML} refers to the general approach aiming at exploiting physical knowledge in order to set up adequate models given the particular data set and the problem. In many cases, models and architectures are, at least partially, enforced by physical laws, such as symmetries or conservation laws \cite{BK,KKLPWY}. Taking this into account dramatically increases efficiency, transparency and robustness of ML algorithms. The very general idea standing behind this approach is the parsimony principle, one of the most universal principles in Science.  

Although physics informed ML is regarded a very recent field, it has been developed upon the blend of ideas from computational physics and energy-based ML. Indeed, concepts of energy and entropy are built in early ML algorithms dealing with problems that are not necessarily related to any physical system \cite{LCHRH}. A classical example of this kind is the famous Hopfield model. We also refer to \cite{HTAL} for energy-based approaches in RL.   

More generally, the term {\it theory informed ML} refers to  architectures which are imposed by a certain theoretical knowledge.

Approaches and models we propose in subsequent sections can be viewed as both physics informed and geometry informed ML. Moreover, many of them are also energy-based models. 

\section{Kuramoto model}

The famous Kuramoto model \cite{Kuramoto} has been introduced in 1975 as a paradigm for the synchronization phenomena in ensembles of coupled oscillators. Following the pioneering Kuramoto's paper, various modifications and generalizations of his model have been proposed. The model describes an ensemble of phase oscillators, whose states are represented by phases $\varphi_i \in [0,2 \pi]$, while amplitudes are neglected.  
 
We consider the model where the dynamics of oscillators are given by the following system of ODE's
\begin{equation}
\label{Kuramoto_123}
\dot \varphi_j(t) = \omega + \frac{1}{N} \sum \limits_{i=1}^N K_{ij} \sin(\varphi_i(t) - \varphi_j(t) - \beta_{ij}) + \xi_j(t), \quad j=1,\dots,N.
\end{equation}
 Here, pairwise interactions are given by the coefficients $K_{ij} \in {\mathbb R}$ and phase shifts $\beta_{ij} \in [0,2 \pi)$. Coefficients $K_{ij}$ represent coupling strengths and each of them can be positive (attractive coupling) or  negative (repulsive coupling). Throughout this paper we will assume that interactions are symmetric, i.e. $K_{ij}=K_{ji}$ for $1 \leq i,j \leq N$. Further, (\ref{Kuramoto_123}) describes the ensemble with noisy frequencies $\omega \in {\mathbb R} + \xi_j(t)$, where $\xi_j(t)$ denote the zero mean Gaussian fluctuations with variance $\kappa>0$ and covariance function $\delta$, that is:
\begin{equation}
\label{Gauss_noise}
\langle \xi_i(t) \rangle = 0; \quad \langle \xi_i(t) \xi_i(t') \rangle = 2 \kappa \delta_{ij} \delta(t-t'),
\end{equation}
where $\delta_{ij}$ stands for the Kronecker delta symbol. 
 


\subsection{Kuramoto models from the geometric point of view}

The states of oscillators in (\ref{Kuramoto_123}) at each moment $t$ are given by their phases $\varphi_j(t)$, that is - by points on the unit circle ${\mathbb S}^1$. 
In order to pass to complex variables, substitute $z_j = e^{i \varphi_j}$. Then an individual phase oscillator evolves by the ODE
\begin{equation}
\label{phase_oscill_indiv}
\dot z_j(t) = i g_j(t) z_j(t), \mbox{  where  } g_j(t) = \omega + \xi_j(t).
\end{equation}
Furthermore, the dynamics of coupled oscillators (\ref{Kuramoto_123}) is rewritten as the system of complex-valued ODE's
\begin{equation}
\label{complex_Riccati}
\dot z_j(t) = i(f_{ij} z_j^2(t) + g_j(t) z_j(t) + \bar f_{ij}), \quad j=1,\dots,N,
\end{equation}
with the coupling functions $f_{ij} \equiv f_{ij}(z_1,\dots,z_N)$ given by 
\begin{equation}
\label{coupling_1}
f_{ij} = \frac{i}{N} \sum \limits_{j=1}^N K_{ij} e^{i (\varphi_j - \beta_{ij})} = \frac{i}{N} \sum \limits_{j=1}^N K_{ij} e^{-i \beta_{ij}} z_j.
\end{equation}

The notation $\bar f_{ij}$ stands for the complex conjugate of a complex number.

Notice that coupling functions in (\ref{complex_Riccati}) do not necessarily need to be of the form (\ref{coupling_1}). One can obtain variations of the Kuramoto model by introducing different effects into the coupling functions $f_{ij}$. For instance, one can include time delays or noise. In any case, dynamics (\ref{complex_Riccati}) preserves the unit circle, that is - $z_j(0) \in {\mathbb S}^1$ implies $z_j(t) \in {\mathbb S}^1$ for all $t>0$.

\begin{remark}
The general Kuramoto model (\ref{complex_Riccati}) includes a broad class of coupled phase oscillators, but many ensembles of phase oscillators are beyond this model. Model (\ref{complex_Riccati}) assumes that oscillators are coupled through the first harmonics only, and not through higher harmonics. For instance, the system with coupling $\sin(\varphi_i - \varphi_j) - \sin 2(\varphi_i - \varphi_j)$ does not belong to the class (\ref{complex_Riccati}). Over the decades the terminology has been mainly established that all the models that can be written in the form (\ref{complex_Riccati}) are referred to as {\it Kuramoto models}. This includes Kuramoto-Sakaguchi, Kuramoto-Daido models, etc. 
\end{remark}

\begin{remark}
From geometric point of view, equations (\ref{complex_Riccati}) can be regarded as Riccati ODE's in the complex plane. Strictly speaking, they are more general than what is traditionally meant by the Riccati ODE, because coefficients $f_{ij}$ are not constant. However, they still preserve cross ratios, which is the characteristic property of Riccati ODE's. Therefore, we will refer to (\ref{complex_Riccati}) as the system of geometric Riccati ODE's on the unit circle.
\end{remark}

\subsection{Hyperbolic geometry of Kuramoto ensembles}

In 1994. Watanabe and Strogatz \cite{WS} demonstrated that the simple Kuramoto ensemble with globally coupled identical oscillators exhibits 3-dimensional dynamics. They have shown that dynamics of a large ensemble can be reduced to the system of ODE's for three global variables. This implies that an ensemble consisting of $N$ oscillators admits $N-3$ constants of motion. This result initiated the new research direction which deals with symmetries and invariant submanifolds in simple Kuramoto networks. 

The underlying symmetries have been exposed in 2009. by Marvel, Mirrolo and Strogatz \cite{MMS}.

In order to explain this, introduce the simplest Kuramoto model, by setting the global (all-to-all) coupling $f_{ij} \equiv f$ in (\ref{complex_Riccati}):
\begin{equation}
\label{Riccati_global}
\dot z_j = i(f z_j^2 + \omega z_j + \bar f), \quad j=1,\dots,N. 
\end{equation} 
The particular choice of the coupling function
$$
f = \frac{i K}{2 N} \sum \limits_{j=1}^N e^{-i (\varphi_j-\beta)}.
$$
yields
\begin{equation}
\label{Kuramoto2}
\dot \varphi_j = \omega + \frac{K}{N} \sum \limits_{i=1}^N \sin(\varphi_i - \varphi_j - \beta), \quad j=1,\dots,N. 
\end{equation}
Denote by ${\mathbb G}$ the group of M\" obius transformations (conformal mappings) in the complex plane which preserve the unit disc ${\mathbb B}^2$. Recall that ${\mathbb G}$ is isomorphic to the matrix group $PSU(1,1) = SU(1,1) / \pm I$ and also to the 3-dimensional Lorentz group $SO^+(2,1)$. We will refer to this group as {\it M\" obius group}.

\begin{proposition} \cite{MMS}

Consider the system (\ref{Riccati_global}) with an arbitrary complex-valued coupling function $f = f(z_1,\dots,z_N)$. There exists a one-parametric family $g_t \in {\mathbb G}$, such that
$$
z_j(t) = g_t(z_j(0)), \quad j=1,\dots,N, \quad \forall t > 0.
$$
\end{proposition}

Proposition 1 has several enlightening corollaries. For instance, it follows that the system (\ref{Riccati_global}) can be restricted to 3-dimensional dynamics on the group ${\mathbb G}$. In other words, system (\ref{Riccati_global}) generates a trajectory in ${\mathbb G}$. Notice that this trajectory depends on initial phases of oscillators $z_1(0),\dots,z_N(0)$. Furthermore, Proposition 1 explains constants of motion in the system (\ref{Riccati_global}): there are $N-3$ constants of motion which correspond to $N-3$ linearly independent cross ratios of 4-tuples of points $z_j(t)$.

Further insights into the relation between hyperbolic geometry and Kuramoto models have been reported in \cite{CEM}. It has been demonstrated that (under certain conditions on the coupling function $f$) Kuramoto dynamics of the form (\ref{Riccati_global}) are gradient flows in the unit disc with respect to hyperbolic metric. Potential function for dynamics (\ref{Kuramoto2}) has particularly transparent geometric interpretation. It turns out that dynamics of Kuramoto ensembles with repulsive interactions uncover a point inside the unit disc that has the minimal sum of hyperbolic distances to the initial points on ${\mathbb S}^1$. In complex analysis this point is conformal barycenter \cite{DE} of the initial configuration.

\subsection{Kuramoto models with several globally coupled sub-ensembles}

Further, consider an ensemble consisting of $D$ sub-ensembles, each of them globally coupled, with inter-ensemble couplings differing from intra-ensemble ones. Let $N_1,\dots,N_D$ be numbers of oscillators in each sub-ensemble. The dynamics is governed by the following system
\begin{equation}
\label{Kuramoto_sub-ensemble}
\dot \varphi_j^l = \omega + \sum \limits_{k=1}^D \frac{K_{kl}}{N_k} \sum \limits_{i=1}^{N_k} \sin(\varphi_i^k - \varphi_j^l - \beta_{kl}), \quad j=1,\dots,N_l, \; l=1,\dots,D
\end{equation}

It has been shown that systems of the form (\ref{Kuramoto_sub-ensemble}) with only $D=2$ display unexpected and highly non-trivial temporal patterns \cite{HS,AMSW}. 

Proposition 1 can be generalized to the system (\ref{Kuramoto_sub-ensemble}) as follows:

\begin{corollary}

Consider the system (\ref{Kuramoto_sub-ensemble}) and denote $z_j^l(t) = e^{i \varphi_j^l(t)}$. There exist one-parametric families $g_t^1, \dots, g_t^D \in {\mathbb G}$, such that
$$
z_j^l(t) = g_t^l(z_j^l(0)), \quad j=1,\dots,N_l, \quad l=1,\dots,D, \quad \forall t > 0.
$$
\end{corollary}

In other words, each sub-ensemble evolves by actions of the M\" obius group ${\mathbb G}$. Moreover, these actions are coupled, as collective motions of one sub-ensemble depend on distributions of oscillators from all the other sub-ensembles. In other words, the system (\ref{Kuramoto_sub-ensemble}) generates a trajectory in the group ${\mathbb G} \times \cdots \times {\mathbb G}$. This trajectory depends on the system parameters $K_{kl}$ and $\beta_{kl}$ as well as on initial positions of all oscillators. 

\section{Kuramoto models on higher-dimensional manifolds}

In the last two decades there have been systematic investigations of dynamical systems describing collective motions of interacting particles on manifolds. Models of this kind have been introduced and studied in various fields of Science and Engineering with different goals. These seemingly unrelated research efforts investigated essentially the same class of dynamical systems. Most of them can be treated as generalizations of the classical Kuramoto models to higher-dimensional manifolds. 

The state space of a Kuramoto oscillator is the unit circle. The first class of extensions is inspired by the fact that the circle is one-dimensional sphere. Hence, many researchers considered generalized oscillators evolving on the sphere ${\mathbb S}^d$ and introduced the system of equations for coupled generalized oscillators \cite{CGO,LMS,CLP,JC-NHM}. Second, circle is a Lie group. This observation led to generalizations of the Kuramoto model to higher-dimensional Lie groups, where frequencies belong to the corresponding Lie algebras \cite{Lohe2009,HDW}. Third, the circle is homeomorphic to the real projective line ${\mathbb R}P^1$, which is the simplest Grassmannian manifold. Hence, consensus gradient algorithms on Grassmannians \cite{Sarlette,SS} can also be regarded as generalizations of the Kuramoto model. Finally, Kuramoto models have also been extended to Stiefel manifolds \cite{MTG} and some other homogeneous spaces \cite{Sarlette}.

From the mathematical point of view, classical Kuramoto models are systems of geometric Riccati ODE's in the complex plane (or on matrix group $SO(2)$). All generalizations to higher-dimensional manifolds are systems of geometric Riccati matrix ODE's \cite{Lohe,Zelikin} (see Remark 2).

\subsection{Non-Abelian Kuramoto models on Lie groups}

State of a Kuramoto oscillator at each moment $t>0$ is represented by a point on the unit circle. The circle ${\mathbb S}^1$ is a group manifold of the Lie group $SO(2) \simeq U(1)$. Then frequencies are elements of the corresponding Lie algebra $\mathfrak{so}(2) \simeq i {\mathbb R}$. From this point of view, (\ref{Kuramoto_123}) is the dynamical system on the $N$-product $SO(2) \times \cdots \times SO(2)$.

The above point of view on classical Kuramoto models indicates potential extensions to higher-dimensional Lie groups. Conceive an abstract particle whose state at each moment $t \geq 0$ is represented by an element of a Lie group ${\mathbb M}$. Suppose that this particle has an intrinsic frequency which belongs to the corresponding Lie algebra ${\mathfrak m}$. Motions of an individual particle on the group ${\mathbb M}$ are given by the following linear ODE:
$$
\dot p_j = b_j p_j, \mbox{  where  } p_j \in {\mathbb M}, \; b_j \in {\mathfrak m}.
$$
The above ODE extends the notion of classical phase oscillator as it turns into (\ref{phase_oscill_indiv}) for the particular case ${\mathbb M} = SO(2)$.

Systems describing collective motions of generalized oscillators on matrix groups have been introduced by Lohe \cite{Lohe2009,Lohe}. The most relevant for our goals are models on special orthogonal groups $SO(n)$ and on unitary groups $U(d)$.

We first introduce the model on the Lie group $U(d)$. The corresponding Lie algebra ${\mathfrak u}(d)$ consists of $d \times d$ skew-Hermitian matrices. Hence, generalized frequencies are elements of this algebra. Dynamics of an ensemble of generalized coupled oscillators on $U(d)$ reads
\begin{equation}
\label{Kuramoto_U(d)}
i \dot U_j U_j^* = H - \frac{i}{2N} \sum \limits_{i=1}^N K_{ij} (U_j U_i^* - U_i U_j^*), \quad j=1,\dots,N
\end{equation}
where $H$ is a skew-Hermitian matrix and the notion $U_j^*$ stands for the complex conjugate of $U_j$. The interaction network is given by the coefficients $K_{ij}$ which can be negative. 

Suppose that the initial conditions for the dynamics (\ref{Kuramoto_U(d)}) satisfy $U_1(0),\dots,U_N(0) \in U(d)$. Then at each moment $t>0$ matrices $U_j(t)$ are unitary. In other words, dynamics (\ref{Kuramoto_U(d)}) are restricted on the group $U(d)$.

Dynamics (\ref{Kuramoto_U(d)}) for $d=1$ reduce to the classical Kuramoto model (\ref{Kuramoto_123}) with $\beta_{ij} = 0$ and $\xi_j = 0$.

Further, consider the Lie group $SO(n)$. The corresponding Lie algebra ${\mathfrak so}(n)$ consists of anti-symmetric matrices. Kuramoto model on $SO(n)$ reads
\begin{equation}
\label{Kuramoto_SO(n)}
\dot Q_j Q_j^T = J + \frac{1}{2N} \sum \limits_{i=1}^N K_{ij} (Q_i Q_j^T - Q_j Q_i^T), \quad j=1,\dots,N
\end{equation}
where $J$ is anti-symmetric matrix and the notion $Q_i^T$ stands for the transpose of $Q_i$ and $K_{ij} \in {\mathbb R}$ are (possibly negative) coupling strengths between (generalized) oscillators $i$ and $j$.
Then at each moment $t>0$ $Q_j(t)$ are special orthogonal matrices, i.e. elements of $SO(n)$. 

Again, (\ref{Kuramoto_SO(n)}) for $n=2$ reduces to the particular case of the Kuramoto model (\ref{Kuramoto_123}) with $\beta_{ij} = 0$ and $\xi_j = 0$.

\subsection{Kuramoto models on spheres}

Classical Kuramoto models can also be generalized to spheres. It turns out that there exist two non-equivalent ways to do that.

\subsubsection{Real Kuramoto models on spheres}

Consider an abstract particle evolving on the unit sphere ${\mathbb S}^{d-1}$ in $d$-dimensional vector space by the following ODE:
\begin{equation}
\label{indiv_oscill_real_sphere}
\dot x = A x, \mbox{  where  } x \in {\mathbb S}^{d-1}, \quad A \in {\mathbb R}^{d \times d}, \; A = - A^T.
\end{equation}
The equation (\ref{phase_oscill_indiv}) describing the motions of an individual phase oscillator is a particular case of (\ref{indiv_oscill_real_sphere}) for $d=2$.

Now, consider an ensemble of coupled generalized Kuramoto oscillators evolving on the sphere satisfying the following system \cite{LMS}
\begin{equation}
\label{Kuramoto_sphere_real}
\dot x_j = A x_j + f_j - \langle x_j, f_j \rangle x_j, \quad j=1\dots,N. 
\end{equation}
Here, $x_j$ are unit vectors in the $d$-dimensional vector space. The notion $\langle \cdot,\cdot \rangle$ stands for the inner product in ${\mathbb R}^d$. Coupling functions $f_j \equiv f_j(x_1,\dots,x_N)$ take values in ${\mathbb R}^d$. It is easy to verify that $x_1(0),\dots,x_N(0) \in {\mathbb S}^{d-1}$ implies that $x_1(t),\dots,x_N(t) \in {\mathbb S}^{d-1}$ for all $t>0$. Therefore, the sphere ${\mathbb S}^{d-1}$ is invariant for dynamics (\ref{Kuramoto_sphere_real}).

\subsubsection{Complex Kuramoto models on spheres}

There exists another non-equivalent way to generalize Kuramoto models to spheres in even-dimensional vector spaces. Suppose that $d=2m$ and represent oscillators as unit vectors in the complex vector space ${\mathbb C}^m$. Consider an individual oscillator evolving by the following linear ODE:
$$
\dot \zeta = H \zeta, 
$$
where $\zeta \in {\mathbb S}^{2m-1} \subset {\mathbb C}^m$ and $H$ is a $m \times m$ anti-Hermitian matrix, interpreted as intrinsic frequency of the generalized oscillator. 
 
Dynamics of an ensemble of generalized Kuramoto oscillators on the sphere in the complex vector space ${\mathbb C}^m$ reads \cite{LMS,Tanaka}
\begin{equation}
\label{Kuramoto_sphere_complex}
\dot \zeta_j = H \zeta_j + g_j - \langle \zeta_j,g_j \rangle_{\mathbb C} \zeta_j, \quad j=1,\dots,N. 
\end{equation}
Here, $\zeta_j(t)$ are unit vectors in ${\mathbb C}^m$ and $\langle \cdot,\cdot \rangle_{\mathbb C}$ denotes an inner product in the $m$-dimensional complex vector space. $g_j \equiv g_j(\xi_1,\dots,\xi_N)$ are coupling functions taking values in ${\mathbb C}^m$.

\begin{remark}
It has been shown \cite{LMS} that models (\ref{Kuramoto_sphere_real}) and (\ref{Kuramoto_sphere_complex}) are not equivalent for $d=2m>2$.  Hence, there are two non-equivalent generalizations of the Kuramoto model to spheres ${\mathbb S}^3,{\mathbb S}^5,{\mathbb S}^7,\dots$. 

Exceptionally, in the case $d=2m=2$, both systems reduce to the classical Kuramoto model (\ref{complex_Riccati}) on ${\mathbb S}^1$.
\end{remark}

Models (\ref{Kuramoto_sphere_real}) and (\ref{Kuramoto_sphere_complex}) are referred to as real and complex Kuramoto models on spheres, respectively.

\begin{remark}
Notice that (\ref{Kuramoto_sphere_real}) and (\ref{Kuramoto_sphere_complex}) assume general coupling functions $f_j$ and $g_j$, respectively. In general, these coupling functions may include the phase shift, time-delay, etc, thus giving rise to different models \cite{JC-NHM}. The standard coupling functions
\begin{equation}
\label{coupling_sphere_real}
f_j = \frac{1}{N} \sum \limits_{i=1}^N K_{ij} x_i 
\end{equation}
and
\begin{equation}
\label{coupling_sphere_complex}
g_j = \frac{1}{N} \sum \limits_{i=1}^N K_{ij} \zeta_i
\end{equation}
yield standard real and complex models on spheres, respectively \cite{CJM}.
\end{remark}

\subsubsection{Hyperbolic geometry of Kuramoto models on spheres}

Recently, the relation between Kuramoto models on spheres and hyperbolic geometries of unit balls has been exposed \cite{LMS}. In order to explain this, consider the simplest setup: ensembles of identical oscillators with global coupling. Then systems (\ref{Kuramoto_sphere_real}) and (\ref{Kuramoto_sphere_complex}) are respectively rewritten as
\begin{equation}
\label{Kuramoto_sphere_real_global}
\dot x_j = A x_j + f - \langle x_j, f \rangle x_j, \quad j=1\dots,N.
\end{equation}
and
\begin{equation}
\label{Kuramoto_sphere_complex_global}
\dot \zeta_j = H \zeta_j + g - \langle \zeta_j,g \rangle_{\mathbb C} \zeta_j, \quad j=1,\dots,N. 
\end{equation}
Further, denote by ${\mathbb Q}$ the group of conformal mappings of the unit ball ${\mathbb B}^d$. ${\mathbb Q}$ is the group of isometries of ${\mathbb B}^d$ in hyperbolic metric \cite{Stoll,Parker}. This group is isomorphic to $SO^+(d,1)$ with the dimension $d(d+1)/2$. It turns out that oscillators in the model (\ref{Kuramoto_sphere_real_global}) evolve by actions of ${\mathbb Q}$.

\begin{proposition} \cite{LMS}

Let oscillators $x_j(t)$ evolve by (\ref{Kuramoto_sphere_real_global}). Then there exists a one-parametric family of hyperbolic isometries $q_t \in {\mathbb Q}$, such that 
$$
x_j(t) = q_t(x_j(0)), \quad \forall t>0, \quad j=1,\dots,N.
$$
\end{proposition}

On the other hand, models (\ref{Kuramoto_sphere_complex_global}) are related to another geometry in unit balls. Denote by ${\mathbb K}$ the group of isometries of ${\mathbb B}^d$ in the Bergman metric \cite{Stoll,Parker}. 

\begin{proposition} \cite{LMS}

Let oscillators $\zeta_j(t)$ evolve by (\ref{Kuramoto_sphere_complex_global}). Then there exists a one-parametric family of Bergman isometries $k_t \in {\mathbb K}$, such that 
$$
\zeta_j(t) = k_t(\zeta_j(0)), \quad \forall t>0, \quad j=1,\dots,N.
$$
\end{proposition}

Propositions 2 and 3 assert that systems (\ref{Kuramoto_sphere_real_global}) and (\ref{Kuramoto_sphere_complex_global}) generate trajectories in groups ${\mathbb Q}$ and ${\mathbb K}$, respectively. These trajectories are determined by the initial distributions of oscillators.

Moreover, a broad class of systems (\ref{Kuramoto_sphere_real_global}) with global coupling are gradient flows in the unit ball in hyperbolic metric \cite{LMS}. The potential function for these gradient flows is particularly transparent for the case of standard coupling function 
\begin{equation}
\label{coupling_standard_sphere}
f_j \equiv f = \frac{K}{N} \sum \limits_{i=1}^N x_i.
\end{equation}
We refer to \cite{LMS} for more details on these hyperbolic gradient flows.

\subsection{Kuramoto models on spheres with several globally coupled sub-ensembles}

In subsection 3.3 we introduced Kuramoto models (\ref{Kuramoto_sub-ensemble}) with several sub-ensembles, each of them globally coupled.
Analogous models can be introduced on higher-dimensional manifolds as well. For our goals, the most important are models on spheres. Here, we consider the real Kuramoto model of this kind, consisting of $D$ sub-ensembles, with $N_k$ oscillators in the $k$-th sub-ensemble:
\begin{equation}
\label{Kuramoto_sub_ensemble_sphere}
\dot x_j^l = A x_j^l + f^l  - \langle x_j^l, f^l \rangle x_j^l,  \quad j=1\dots,N_l,\; k,l=1,\dots,D
\end{equation}
where $x_j^l \in {\mathbb S}^{d-1}$ and $f^{kl} \equiv f^{kl}(x_1^k,\dots,x_{N_k}^k,x_1^l,\dots,x_{N_l}^l)$ is the coupling functions between sub-ensembles indexed by $k$ and $l$.

For instance, one can suppose that coupling functions $f^{kl}$ are of the following form
\begin{equation}
\label{coupling_sub-ensemble_sphere}
f^l = \frac{1}{D} \sum \limits_{i=1}^D \frac{K_{il}}{N_i} \sum \limits_{j=1}^{N_i} Q_{il} x_j^i.
\end{equation}
Here, $Q_{il}$ are $SO(d)$ matrices, generalizing the notion of phase-shifted coupling.

The slight generalization of Proposition 2 yields:

\begin{corollary}

Let oscillators $x_j^l(t)$ evolve by (\ref{Kuramoto_sub_ensemble_sphere}). Then there exist one-parametric families $q^1_t,\dots,q_t^D \in {\mathbb Q}$, such that 
$$
x_j^l(t) = q_t^l(x_j(0)), \quad \forall t>0, \quad j=1,\dots,N_l, \; l=1,\dots,D.
$$
\end{corollary}

The above corollary asserts that (\ref{Kuramoto_sub_ensemble_sphere}) generates a trajectory in the $D$-product ${\mathbb Q} \times \cdots \times {\mathbb Q}$. This trajectory depends on the initial distribution of all oscillators.

\subsection{Kuramoto models as gradient flows}

One favorable property of Kuramoto models is that many of them exhibit potential dynamics. As a rule, ensembles with identical oscillators and symmetric couplings are gradient flows (sometimes in more than one sense).

The system (\ref{Kuramoto_SO(n)}) with zero frequencies $J=0$ is gradient flow for the potential function
\begin{equation}
\label{potential1}
V(Q_1,\dots,Q_N) = \frac{1}{2N^2} \sum \limits_{i=1}^N \sum \limits_{j=1}^N K_{ij} Tr(Q_i^T Q_j), 
\end{equation}
where $Tr$ denotes the trace.

For the classical Kuramoto model (on the group $SO(2)$) (\ref{potential1}) reduces to
$$
V_K(\varphi_1,\dots,\varphi_N) = \frac{1}{2N^2} \sum \limits_{i=1}^N \sum \limits_{j=1}^N K_{ij} \sin^2 \frac{\varphi_j-\varphi_i}{2}.
$$

Analogously, (\ref{Kuramoto_U(d)}) with zero frequencies $H=0$ is the gradient flow for
\begin{equation}
\label{potential2}
V(U_1,\dots,U_N) = \frac{1}{2N^2} \sum \limits_{i=1}^N \sum \limits_{j=1}^N K_{ij} Tr(U_i^* U_j). 
\end{equation}

Kuramoto models on spheres with zero frequencies and symmetric coupling also exhibit potential dynamics. The model (\ref{Kuramoto_sphere_real}) with the coupling function (\ref{coupling_sphere_real}) is gradient flow for the potential
\begin{equation}
\label{potential_sphere}
V_{sph}(x_1,\dots,x_N) = \frac{1}{2N^2} \sum \limits_{i=1}^N \sum \limits_{j=1}^N K_{ij} |x_i-x_j|^2 = \frac{1}{2 N^2} \sum \limits_{i=1}^N \sum \limits_{j=1}^N K_{ij} (1-\cos \theta_{ij}).
\end{equation}
Here, $\theta_{ij}$ denotes the angle between $x_i$ and $x_j$.

Underline that models on spheres, as well as on special orthogonal and unitary groups are gradient flows in the chordal metric on these manifolds \cite{Sepulchre}. On spheres, this is equivalent to the cosine metric.

Consensus algorithms on Grassmannian manifolds are gradient flows as well, we refer to \cite{Sarlette} for an explanation.

The above observations further imply that by adding the noise in an appropriate way to the above systems we obtain various Langevin dynamics.  
For instance, (\ref{Kuramoto_123}) are Langevin dynamics for the potential
\begin{equation}
\label{energy}
E(\varphi_1,\dots,\varphi_N) = \frac{1}{2N^2} \sum \limits_{i=1}^N \sum \limits_{j=1}^N K_{ij} \cos(\varphi_i - \varphi_j - \beta_{ij}).
\end{equation}

As already noted above, recent geometric investigations of Kuramoto models have shown that models on spheres (\ref{Kuramoto_sphere_real}) with global coupling induce gradient flows in hyperbolic balls \cite{LMS}. In general, the question of hyperbolic gradient flows induced by Kuramoto models on spheres is still to be fully explored. In particular, it is interesting to examine conditions for the potential dynamics on hyperbolic multi-discs and multi-balls which are induced by the models with several sub-ensembles. The second interesting question is adding the noise in an appropriate way in order to obtain Langevin dynamics on hyperbolic balls. 

Underline that potential functions for hyperbolic gradient flows depend on initial positions of oscillators on ${\mathbb S}^{d-1}$.  Hence, learning potentials in hyperbolic balls involves learning initial configuration of oscillators on a sphere. On the other side, learning potentials (\ref{potential1},\ref{potential2},\ref{potential_sphere}), boils down to learning the interaction network $K_{ij}$.   

\subsection{Consensus algorithms on other manifolds}

As exposed in the previous subsection, non-Abelian Kuramoto models with equal frequencies exhibit potential dynamics. Potential functions (\ref{potential1}) and (\ref{potential2}) measure the total disagreement between generalized oscillators. Therefore, collective dynamics of generalized Kuramoto oscillators with all positive couplings can be regarded as continuous-time consensus algorithms minimizing the total disagreement. If all couplings are negative, the term anti-consensus is used.

The same is valid for Kuramoto models on spheres and their disagreement function (\ref{potential_sphere}).

Geometric consensus theory investigates consensus algorithms on various homogeneous manifolds \cite{Sarlette}. In addition to particular manifolds discussed above, potentially the most interesting are the Grassmannians. Matrix Riccati ODE's generate flows on Grassmaniann manifolds $G_p({\mathbb R}^n)$. Indeed, if a subspace $W$ evolves by linear Hamiltonian system, then the local coordinates of $W$ on $G_p({\mathbb R}^n)$ satisfy the matrix Riccati ODE \cite{Zelikin}. 

Moreover, the Riccati flow on $G_n({\mathbb R}^{2n})$ is closely related to actions of the group $SL(2n,{\mathbb R})$. One can derive the matrix Riccati ODE by taking a linear combination of infinitesimal generators of $SL(2n,{\mathbb R})$. For the case $n=1$ this has been used for the geometric proof that identical globally coupled Kuramoto oscillators evolve by actions of the M\" obius group (see Proposition 1 above). Recall that the M\" obius group $PSU(1,1)$ is isomorphic to $SL(2,{\mathbb R})$. 

Riccati flows on Grassmannian manifolds have been applied to computing an invariant subspace of a matrix and to averaging subspaces \cite{AMS1}. System of ODE's describing the gradient flow (consensus algorithm) on $G_p({\mathbb R}^n)$ has been discussed in \cite{Sarlette}.
 
Finally, we mention that consensus algorithms have also been introduced on Stiefel manifolds \cite{MTG}, while Riccati flows have been studied on Siegel domains as well \cite{Zelikin}.

\section{Directional statistics and swarms on manifolds for probabilistic modeling and inference on Riemannian manifolds}

Many ML algorithms involve stochastic policies over finite sets of outcomes. In such cases the statistical model is provided by the family of categorical distributions (i.e. family of probability distributions over a finite set of outcomes). This family is isomorphic to the unit simplex
$$
\triangle^D = \{ p_1,\dots,p_D \; : \; p_1+\cdots+p_D = 1, \, p_1 \geq 0, \dots, p_D \geq 0 \}.
$$ 
In the absence of some case-specific information, the natural choice for the prior is the distribution where all probabilities are equal: $p_1 = \cdots = p_D = 1/D.$ Clearly, this is the maximum entropy distribution in this family.  

When it comes to problems over continuous spaces, the most frequently exploited statistical model is, of course, the Gaussian family ${\cal N}(a,\Sigma)$. This family is parametrized by the mean vector $a \in {\mathbb R}^n$ and positively defined covariance matrix $\Sigma \in {\mathbb R}^{n \times n}$. A great deal of algorithms in supervised ML involve some sort of optimization over ${\cal N}(a,\Sigma)$. Furthermore, stochastic policies over continuous action spaces in RL are typically modeled by Gaussians. The Gaussian family is also widely used in the black-box optimization. The most prominent stochastic search algorithm, named Covariance Matrix Adaptation - Evolution Strategy (CMA ES) \cite{OAAH} performs an update over ${\cal N}(a,\Sigma)$.

In those cases when representative power of the Gaussian family is insufficient, the mixtures of Gaussians provide the most popular option.

Non-Gaussian probability distributions are rarely used in ML. Even when dealing with inference problems over compact subsets of ${\mathbb R}^n$, researchers prefer to use the Gaussian statistical model, with a very few exceptions \cite{CMS}.

Such a domination of the Gaussian family has a strong justification. Indeed, this family posses a combination of advantages which render it a universal model in probabilistic ML. The only potential alternative is the family of multivariate Cauchy distributions. However, in most cases this family is immediately ruled out due to the fact that Cauchy distributions do not have finite moments. We list four favorable properties of the family ${\cal N}(a,\Sigma)$:

P1) Given the fixed mean vector and the covariance matrix, Gaussian distributions are the maximum entropy distributions. Hence, they are natural choices for priors from the Bayesian point of view \cite{Jeffreys}. 

P2) The Gaussian family belongs to the class of exponential families of probability distributions, which implies all the desirable properties concerning reparametrization, sufficient statistics and duality in convex optimization (Legendre-Fenchel transform) \cite{Barndorff-Nielsen}.

P3) The Fisher information metric for the Gaussian family has a simple closed-form expression. This property enables the design of efficient and reasonably simple algorithms that follow the natural gradient update \cite{Amari,MP1,Martens}. In particular, this property provides a theoretical justification for Natural Evolution Strategies in the black-box optimization \cite{OAAH}.

P4) The Gaussian family is invariant with respect to the group of affine transformations in ${\mathbb R}^d$.

With recent advances in Geometric Deep Learning, there is a rising number of tasks which require probabilistic modeling and inference over non-Euclidean spaces. When dealing with problems of this kind, ignoring intrinsic geometry of the data leads to incorrect or inaccurate algorithms. In such setups the Gaussian family becomes an inappropriate choice. Instead, one needs suitable and tractable statistical models over Riemannian manifolds. This fact has been widely recognized only recently and investigations in this direction are at a very early stage. Future advances will rely on results of directional statistics.  

The family satisfying all properties (P1)-(P4) does not exist on curved spaces. Having that in mind, the key issue consists in determining which of these properties are the most essential for ML.

Property (P1) is not very significant in practice. On compact manifolds the uniform distribution posses the MaxEnt property and provides an obvious choice for the prior. Therefore, it is necessary to work with those statistical models which contain the uniform distribution.

On the other hand, property (P4) plays an essential role in many ML algorithms which employ the Gaussian family. Therefore, it is highly desirable to deal with families of probability distributions on spheres, tori and homogeneous spaces that exhibit the group-invariance property analogous to (P4).

This Section contains an overview of families of probability distributions over Riemannian manifolds that have been (or should be) used in ML. The overview is accompanied with a brief discussion on advantages and drawbacks of the corresponding statistical models. The main point of the present section is that some of the most suitable statistical models on curved spaces have been neglected in ML so far. 

We will point out those distributions which are associated with Kuramoto and swarming dynamics in the continuum limit, i.e. by passing to the continuum of oscillators or particles. Then dynamics are governed by evolution equations (first order PDE's on manifolds). In such cases, symmetries impose low-dimensional evolution of densities on invariant submanifolds.

In the presence of noise, continuum limit yields the second-order PDE involving the diffusion term (the Fokker-Planck equation). We will be interested in distributions which can be generated by the swarming dynamics, either as stationary distributions in noisy models in the long time, or as invariant families in deterministic (noiseless) dynamics. Notice that the later satisfy the group-invariance property (P4).

In such a way we identify models that are potentially relevant for the probabilistic-geometric ML. At the same time, we systematize some mutually related recent advances in directional statistics and physics of complex systems. We refer to \cite{PG-P} for a more comprehensive overview of probability distributions over circular, toroidal and cylindrical geometries. Moreover, \cite{PG-P} also contains the discussion on inference methods on Riemannian manifolds with the extensive list of references.
  
\subsection{Statistical models over circles and tori}

We discuss four families of probability distributions that are potentially relevant for ML. Only one of them (von Mises) have been applied to ML problems so far. Two families (von Mises and wrapped Cauchy) belong to wider classes of probability measures on spheres (see the next subsection). We omit some families on the circle (such as wrapped normal) which we do not find a suitable model in any ML setup. Although inference and density estimation on one-dimensional manifold ${\mathbb S}^1$ may seem as not very challenging problems, algorithms on $d$-dimensional torus $ {\mathbb T}^d = {\mathbb S}^1 \times \cdots \times {\mathbb S}^1$ may have a very high computational cost, if inappropriate models are employed. Therefore, good understanding and adequate choice of statistical models over tori is of a great importance for geometry informed statistical ML. Statistical models for toroidal data are provided by multivariate versions of the models explained below. However, learning covariances in multivariate data is usually complicated with very few mathematical results available. On the other hand, appropriately chosen Kuramoto/swarming models can be trained to learn such covariances even if the exact formulae are not available.     

\subsubsection{von Mises distribution}

The von Mises probability distributions $vM(\kappa,\mu)$ are defined by the family of unimodal densities on the circle given by:
\begin{equation}
\label{von_Mises}
p_{vM}(\varphi) = \frac{1}{2 \pi {\cal I}_0(\kappa)} exp\{\kappa \cos(\varphi-\mu)\}, \; \varphi \in {\mathbb S}^1,
\end{equation}
where $\mu \in [0,2 \pi)$ and $\kappa \geq 0$ are parameters, representing the mean direction and concentration, respectively. The notion ${\cal I}_0(\cdot)$ stands for the modified Bessel function of the first kind of order zero.

Density (\ref{von_Mises}) is a particular case of the von Mises-Fisher densities on spheres (see below). This is essentially the only family of probability densities over the circle that has been used in ML so far. 

Statistical model (\ref{von_Mises}) has certain advantages. First, it depends on two parameters only, with the parameter space ${\mathbb S}^1 \times {\mathbb R}_+$. 
The second argument in favor of this statistical model is that von Mises distributions are maximum entropy distributions over the circle with fixed mean \cite{MJ}. In other words, the family (\ref{von_Mises}) satisfies property (P1). In addition, it also satisfies (P2) and (P3). Notice that (\ref{von_Mises}) contains the uniform distribution for $\kappa=0$. 

Multivariate von Mises distributions provide a statistical model for probabilistic modeling and inference on tori. This statistical model has been applied in bioinformatics \cite{MHTS}.

On the other hand, the von Mises family does not satisfy any group-theoretic property analogous to (P4). This drawback renders it inconvenient choice in some architectures and setups.
 
The von Mises family is associated with Kuramoto models of the following form
\begin{equation}
\label{Kuramoto_von Mises}
\dot \varphi_j = \omega + \frac{K}{N} \sum \limits_{i=1}^N \sin(\varphi_i - \varphi_j - \beta) + \xi_j , \quad j=1,\dots,N. 
\end{equation}
where $\xi_j$ is the Gaussian noise defined by (\ref{Gauss_noise}).

By taking limit of the Langevin dynamics (\ref{Kuramoto_von Mises}) for the continuum of oscillators, we obtain the Fokker-Planck PDE for evolution of densities on the circle. This PDE has a stationary density (\ref{von_Mises}).

In conclusion, Kuramoto models with noise can encode von Mises distributions $vM(\kappa,\mu)$ and their mixtures. Algorithm starts from the uniform distribution (corresponding to $\kappa=0$ and arbitrary $\mu$) and performs an update of $\kappa$ and $\mu$ by learning $K$ and $\beta$ in (\ref{Kuramoto_von Mises}). When dealing with torical data, multivariate von Mises distributions can be approximated by adding the noise to models with several sub-ensembles (\ref{Kuramoto_sub-ensemble}).  






\subsubsection{Wrapped Cauchy distribution}

One of the most important families of probability measures on the unit circle is obtained by "wrapping" the Cauchy (Lorentzian) distributions on the real line to the circle. This yields so-called wrapped (or circular) Cauchy distributions with the density functions \cite{McCullagh}
\begin{equation}
\label{wrapped_Cauchy}
p_{wC}(\varphi) = \frac{1}{2 \pi} \frac{1-r^2}{1-2r \cos(\varphi-\Phi) + r^2}, \; \varphi \in {\mathbb S}^1.
\end{equation}

Denote this family by $wC(\alpha)$, where $\alpha = r e^{i \Phi}$ is a point in the unit disc ${\mathbb B}^2$. The Fisher information metric for this family coincides with the hyperbolic metric in the unit disc \cite{CEM}. Hence, the statistical manifold $wC(\alpha)$ is isomorphic to the hyperbolic disc ${\mathbb B}^2$. Therefore, learning over the family $wC(\alpha)$ is equivalent to the optimization over ${\mathbb B}^2$.

Notice that densities (\ref{wrapped_Cauchy}) can be written in another parametrization using the complex variable $z=e^{i \varphi}$ as follows
$$
p_{wC}(z) = \frac{1}{2 \pi} \frac{1 - |\alpha|^2}{|z-\alpha|^2}.
$$

Although $wC(\alpha)$ is one of the central families in directional statistics, it has not been, to our best knowledge, used in ML so far. We argue that $wC(\alpha)$ provides the most convenient and transparent statistical model on circles and tori with many favorable properties. Obviously, densities (\ref{wrapped_Cauchy}) contain a uniform density on ${\mathbb S}^1$ for $r=0$. Furthermore, this family satisfies properties (P2), (P3) and, most importantly, the group-invariance property analogous to (P4). This is substantiated in the following

\begin{proposition}
  
The family $wC(\alpha)$ is invariant with respect to actions of the M\" obius group ${\mathbb G}$. All wrapped Cauchy distributions are obtained as M\" obius transformations of the uniform distribution on the circle.
\end{proposition}

Recalling Proposition 1, we conclude that densities (\ref{wrapped_Cauchy}) can be approximated by the simplest Kuramoto models. Consider the system (\ref{Riccati_global}) in the continuum limit and assume that initial distribution of oscillators is uniform on ${\mathbb S}^1$. Proposition 4 asserts that the density of oscillators at each moment $t$ is of the form (\ref{wrapped_Cauchy}).

Obviously, the representative power of $wC(\alpha)$ is very limited. This statistical manifold is the two-dimensional orbit of the M\" obius group \footnote{In this context we recall negative results on the representative power of NN architectures based on group orbits reported in the famous classical book \cite{MP}.} and consists of unimodal and symmetric densities on the circle. 

Most important, we have a clear understanding of the representative power of architectures associated with this model. One can overcome limitations by using mixtures of $m$ wrapped Cauchy distributions. Such mixtures can be generated by $m$ uncoupled Kuramoto models. 

Family of multivariate wrapped Cauchy distributions provides a suitable model for learning toroidal data, we refer to \cite{Kato,KP} for some theoretical results. Exact formula for the Fisher information is not available on higher-dimensional tori, making it difficult to design accurate algorithms. However, multivariate wrapped Cauchy distributions can be encoded in Kuramoto models with several sub-ensembles (\ref{Kuramoto_sub-ensemble}).

\subsubsection{Kato-Jones distribution}

Kato-Jones distributions constitute the family over ${\mathbb S}^1$ with increased representative power compared to the wrapped Cauchy. This statistical manifold is defined by densities of the form \cite{KJ}
\begin{equation}
\label{Kato-Jones}
p(\varphi) = \frac{1-r^2}{2 \pi {\cal I}_0(\kappa)} exp \left \{ \frac{\kappa \cos(\varphi-\eta) - 2 r \cos \nu}{1+r^2-2r \cos(\varphi-\gamma)} \right \} \cdot \frac{1}{1+r^2-2r \cos(\varphi-\gamma)}, \; 0 \leq \varphi < 2 \pi
\end{equation}
where the parameters are related in the following way: $\gamma = \mu + \nu, \xi = \sqrt{r^4 + 2r^2 \cos(2 \nu) + 1}, \eta = \mu + arg\{ r^2 \cos(2 \nu) + 1 + i r^2 \sin(2 \nu)\}$ with $0 \leq \mu,\nu < 2 \pi, 0 \leq r < 1, \kappa > 0$. Hence, densities (\ref{Kato-Jones}) define the four-dimensional manifold, with the space of parameters ${\mathbb G} \times {\mathbb R}_+$, where ${\mathbb G}$ denotes the M\" obius group, as before.

We denote this family by $KJ(\mu,\nu,r,\kappa)$. It includes the uniform measure on ${\mathbb S}^1$ for $r=0, \kappa=0$. Furthermore, this family is invariant with respect to actions of ${\mathbb G}$. This is substantiated in the following

\begin{proposition}  \cite{KJ}

The family $KJ(\mu,\nu,r,\kappa)$ is invariant with respect to actions of the M\" obius group ${\mathbb G}$. All Kato-Jones distributions with fixed $\kappa$ are obtained as M\" obius transformations of the von Mises $vM(\kappa,0)$ distributions on the circle.
\end{proposition}

Hence, K-J families for fixed $\kappa>0$ are three-dimensional orbits of the M\" obius group.
Applying Proposition 1 we conclude that densities (\ref{Kato-Jones}) appear as a result of the two-stage dynamics. The von Mises distribution $vM(\kappa,0)$ arises as the stationary distribution in the model (\ref{Kuramoto_von Mises}) for $0 \leq t \leq T$ with $T$ sufficiently large. At the second stage dynamics (\ref{Riccati_global}) for $t>T$ transforms von Mises into K-J distributions, we refer \cite{CJ-PLA} for more detailed explanation.

In conclusion, the statistical manifold $KJ(\mu,\nu,r,\kappa)$ is four-dimensional and contains both submanifolds $vM(\kappa,\mu)$ (for $r=0$) and $wC(\alpha)$ (for $\kappa=0$). Representative power of $KJ(\mu,\nu,r,\kappa)$ is higher, as it contains bimodal and asymmetric densities. 

The Fisher information for this family is explicitly calculated\cite{KJ}. Therefore, the K-J family provides a statistical model with improved representative power satisfying properties (P2) and (P4). These distributions have not been used in ML so far. We point out the recent preprint \cite{NKNJ} which works with mixtures of two K-J distributions for the inference problem on daily periodic traffic data.

\subsubsection{Hyperbolic von Mises distribution}

For the last family in this overview, consider the Kuramoto model with the multiplicative noise
\begin{equation}
\label{Kuramoto_hyp von Mises}
\dot \varphi_j = \omega + \frac{K}{N} \sum \limits_{i=1}^N \sin(\varphi_i - \varphi_j - \beta) + \frac{1}{N} \sum \limits_{i=1}^N  \left[\sqrt{1+C \cos(\varphi_i - \varphi_j)}\right] \xi_j , \quad j=1,\dots,N. 
\end{equation}
For $C=0$ (\ref{Kuramoto_hyp von Mises}) reduces to the model (\ref{Kuramoto_von Mises}) with the additive noise. In general (for $C \neq 0$), the stationary distributions for (\ref{Kuramoto_hyp von Mises}) are hyperbolic von Mises \cite{HFB}. Their densities are of the form
\begin{equation}
\label{hyp_von_Mises}
p(\varphi) = \frac{1}{2 \pi P_{-\alpha}^0(\cosh(\eta))cosh^\alpha(\eta)} \left[ 1 + \tanh(\eta) \cos(\psi- \varphi) \right]^{-\alpha}. \; 0 \leq \varphi \leq 2 \pi
\end{equation}
where $P_{-\alpha}^0$ denotes the Legendre polynomial of order zero and degree $-\alpha$.

The parameters of (\ref{hyp_von_Mises}) are $\eta > 0, \alpha \in {\mathbb R}$ and $\psi \in [0,2 \pi)$. The family (\ref{hyp_von_Mises}) contains a uniform distribution for $\eta = 0$.

In conclusion, the Kuramoto model with multiplicative noise (\ref{Kuramoto_hyp von Mises}) can be used to learn over the three-dimensional manifold of hyperbolic von Mises distributions.


\subsection{Statistical models over spheres}

So far, quite a few papers presented ML algorithms based on statistical models over spheres. In the present subsection we point out some recent applications in all three major fields of ML: (un)supervised learning and RL. Majority of applications have been reported for unsupervised learning problems, such as clustering of angular data based on the cosine metric. The paper \cite{GM} contains a brief review of statistical clustering methods on spheres.
 
The von Mises-Fisher family $vMF(\kappa,\mu)$ is the most popular statistical model for ML algorithms over spheres \cite{BDGS,HBMGC,GY,SGM}. The only alternative model used so far is provided by the family of Bingham distributions. In addition, there are a couple of papers experimenting with other (that is - different from von Mises-Fisher and Bingham) options.
 
\subsubsection{von Mises-Fisher distribution}

 The von Mises-Fisher distributions are defined by density functions of the form \cite{MJ}
\begin{equation}
\label{von_Mises-Fisher}
p(x) = C(\kappa) e^{\kappa \mu^T \varphi}, \; x \in {\mathbb S}^{d-1}.
\end{equation}
Here $C(\kappa)$ is the normalization constant, given by
$$
C(\kappa) = \frac{2 \pi^{d/2} {\cal I}_{d/2-1}(\kappa)}{\kappa^{d/2-1}}
$$
where ${\cal I}_\nu(\cdot)$ is the modified Bessel function of order $\nu$ of the first kind.

For important particular case $d=3$ (corresponding to the two-dimensional sphere), the above expression for $C(\kappa)$ simplifies to
$$
C(\kappa) = \frac{\kappa}{2 \pi(e^{\kappa}-e^{-\kappa})}.
$$
Parameters of the family $vMF(\kappa,\mu)$ are concentration $\kappa > 0$ and the mean direction $\mu \in {\mathbb S}^{d-1}$. Hence, the space of parameters is ${\mathbb S}^{d-1} \times {\mathbb R}_+$. The von Mises-Fisher family includes the uniform distribution on ${\mathbb S}^{d-1}$, obtained by setting $\kappa=0$. Densities (\ref{von_Mises-Fisher}) are unimodal and symmetric. 

The family $vMF(\kappa,\mu)$ has been applied to clustering algorithms on spheres \cite{BDGS} and spherical embeddings \cite{SGM}. Along with a small number of parameters to be learned, this family also has an advantage of satisfying properties (P2) and (P3).

Densities (\ref{von_Mises-Fisher}) are stationary for the Fokker-Planck equation (FPE) obtained in the continuum limit of noisy real Kuramoto models on spheres, see \cite{FL,ZTP}. By adding the additive white noise to the system (\ref{Kuramoto_sphere_real_global}), one obtains Langevin dynamics on ${\mathbb S}^{d-1}$. FPE for these dynamics is the Smoluchowski PDE on sphere \cite{FL}. 

Hence, systems of the form (\ref{Kuramoto_sphere_real}) with added Gaussian noise are able to learn over the family $vMF(\kappa,\mu)$. Accordingly, one can generate mixtures of the von Mises-Fisher distributions by solving several uncoupled systems of this kind. Furthermore, multivariate von Mises-Fisher distributions on products of spheres ${\mathbb S}^{d-1} \times \cdots \times {\mathbb S}^{d-1}$ can be learned by noisy Kuramoto models on spheres with several globally coupled sub-ensembles. 

\subsubsection{Spherical Cauchy distribution}

Cauchy family on spheres $sphC(\rho,\mu)$ consists of probability distributions with densities of the form \cite{KMcC}
\begin{equation}
\label{spherical_Cauchy}
P_{hyp}(x) = \frac{\Gamma(\frac{d}{2})}{2 \pi^{d/2}} \left( \frac{1-\rho^2}{1+\rho^2-2 \rho \mu^T x} \right)^{d-1}, \mbox{  where } x \in S^{d-1}.
\end{equation}
The parameters of (\ref{spherical_Cauchy}) are $\mu \in {\mathbb S}^d$ and $0 \leq \rho < 1$. These two parameters can be identified with a point $\zeta \in \rho \cdot \mu$ inside the unit ball. Hence, the parameter space for $sphC(\rho,\mu)$ is the unit ball ${\mathbb B}^d$. It turns out that the natural metric on the $d$-dimensional statistical manifold $sphC(\rho,\mu)$ is (proportional to) hyperbolic metric in ${\mathbb B}^d$ \cite{KMcC}.

The family $sphC(\rho,\mu)$ has not been used in ML algorithms so far, although in our point of view, it provides the most convenient statistical model in most setups. 

First of all, this family satisfies property (P2). The Fisher information has a simple closed form expression which is closely related to hyperbolic metric in the unit ball \cite{KMcC}.

The most important advantage of $sphC(\rho,\mu)$ is the group invariance property similar to (P4). Indeed, this family is invariant for actions of the group of isometries of the unit ball in hyperbolic metric.    

\begin{proposition}
  
The family $sphC(\alpha)$ is invariant with respect to actions of the group ${\mathbb Q}$ of hyperbolic isometries of the unit ball. All spherical Cauchy distributions are obtained by acting with the group ${\mathbb Q}$ on the uniform distribution on the sphere ${\mathbb S}^{d-1}$.
\end{proposition}

From Proposition 6 it follows that the family $sphC(\rho,\mu)$ can be encoded in real Kuramoto models on spheres (\ref{Kuramoto_sphere_real_global}). Indeed, consider the dynamics (\ref{Kuramoto_sphere_real_global}) and assume that the initial distribution of oscillators is uniform on ${\mathbb S}^{d-1}$. Then, due to propositions 2 and 6, the density of oscillators at each moment $t$ is (\ref{spherical_Cauchy}). Therefore, one can train simple Kuramoto models for sampling, modeling and inference problems over $sphC(\rho,\mu)$.

In conclusion, $sphC(\rho,\mu)$ has many favorable properties making it very convenient statistical model for probabilistic modeling and inference on spheres. The simplicity of this model stems from its nice geometric and group-theoretic properties. By taking mixtures of spherical Cauchy, one can achieve satisfactory representative power for most purposes. Real Kuramoto models on spheres with several sub-ensembles can be used for learning multivariate spherical Cauchy as a statistical model on the product ${\mathbb S}^{d-1} \times \cdots \times {\mathbb S}^{d-1}$.

\subsubsection{Bergman-spherical Cauchy distribution}

Another family of distributions on spheres is $BsphC(w)$ given by densities of the form
\begin{equation}
\label{Bergman_spherical_Cauchy}
P_{Berg}(\zeta,w) = \frac{\Gamma(\frac{d=m+1}{2})}{2 \pi^{(m+1)/2}} \frac{(1-|w|^2)^m}{(1-\langle w,\zeta \rangle)^{2m}}, \mbox{  where } \zeta \in S^{2m-1}, \; w \in {\mathbb B}^{2m}.
\end{equation}

To our best knowledge, the family (\ref{Bergman_spherical_Cauchy}) has never been studied in directional statistics, or applied in any field. This family arises as an $m$-dimensional complex statistical manifold, which is invariant with respect to actions of the group ${\mathbb K}$ of Bergman isometries of the unit ball. In other words, while $sphC(\rho,\mu)$ is invariant with respect to the group ${\mathbb Q}$ of isometries of the unit ball in standard hyperbolic metric, $BsphC(w)$ is an invariant submanifold for actions of the group ${\mathbb K}$ of isometries of the unit ball in Bergman metric. We refer to \cite{Stoll} for geometric and harmonic-analytic details on this mathematically sophisticated topic.

Due to Proposition 3, the distributions from $BsphC(w)$ are generated by complex Kuramoto models (\ref{Kuramoto_sphere_complex_global}) on spheres. If the initial distribution of oscillators is uniform on $S^{2m-1}$, then the density of oscillators at each moment $t$ is of the form (\ref{Bergman_spherical_Cauchy}). Hence, complex Kuramoto models on spheres can be trained to learn distributions from $BsphC(w)$.

In conclusion, the family $BsphC(w)$ may seem as an exotic choice for the statistical model in ML problems, but it has significant advantages, as it is low-dimensional and satisfies property (P4). We are not able to discuss advantages and drawbacks of this family compared to $sphC(\rho,\mu)$ in specific setups. This question is related with geometric subtleties and deep understanding of isometries in unit balls.

\begin{remark}
In mathematics, especially in harmonic analysis and potential theory, the notion of Poisson kernel plays a central role. These are integral kernels that appear in solutions of the Dirichlet boundary problems.
\end{remark}

Functions of the form (\ref{spherical_Cauchy}) are {\it Poisson kernels for the hyperbolic Dirichlet problem} \cite{Stoll}. On the other side,(\ref{Bergman_spherical_Cauchy}) are {\it Poisson kernel in the Bergman metric}. To our best knowledge, none of these densities have never been applied in ML so far. 

Two recent studies \cite{GMS,GM} employed {\it Poisson kernels in Euclidean metric} to unsupervised learning (clustering on spheres). We emphasize that (\ref{spherical_Cauchy}) and (\ref{Bergman_spherical_Cauchy}) are more suitable models for dealing with hyperbolic data.

\subsubsection{Bingham distribution}

Some recent papers \cite{JA,GSSAKR} experimented with the Bingham family $Bing(M,Z)$ for RL problems on spheres and rotation groups. This family is defined by densities of the form
\begin{equation}
\label{Bingham}
p(x) = C(Z) exp \left \{ x^T MZM^T x \right \}, \quad x \in {\mathbb S}^{d-1}.
\end{equation}

Parameters of (\ref{Bingham}) are orthogonal $d \times d$ matrix $M$ and diagonal $d \times d$ matrix $Z$.

The normalization constant in (\ref{Bingham}) is given by
$$
C(Z) = \; _1F(\frac{1}{2};\frac{n}{2};Z)^{-1},
$$
where $_1F(\cdot;\cdot;\cdot)$ stands for confluent hypergeometric series.

Notice that (\ref{Bingham}) contains many parameters. Unlike two previous families, this family does not posses nice group-theoretic properties. Hence, the choice of this statistical model entails much more involved estimation of parameters.

A characteristic property of Bingham distributions is that they are antipodally symmetric. This property is advantageous when dealing with rotations in the three-dimensional space (statistical learning over the group $SO(3)$), because it naturally avoids so-called {\it gimbal loop} (a well known error caused by the fact that two antipodal points on ${\mathbb S}^3$ correspond to the same 3D rotation). 

\subsection{Statistical models over hyperbolic spaces}

Although probability measures over hyperbolic spaces may seem an exotic topic at the first glance, recent developments in Geometric DL indicate that well understood and tractable families of this kind would play a central role in many ML algorithms. Since many data sets coming from different fields are naturally embedded into hyperbolic spaces, the corresponding statistical models are necessary for sampling, density estimation, inference and probabilistic modeling.

The most tractable probability distribution of this kind appears in \cite{KPKVB} with applications to hyperbolic embeddings and modeling complex networks. The suggested distribution over the unit hyperbolic disc ${\mathbb B}^2$ is obtained by sampling an angle $\theta$ from the uniform distribution over $[0,2 \pi]$ and radius $r$ from the distribution over $[0,1]$ defined by the density
$$
p(r) = \frac{2 \sinh r}{e + e^{-1} - 2}.
$$
Then the random point $\zeta = r \theta \in {\mathbb B}^2$.

Another possible approach is based on the extension of densities from the unit circle to disc, using the notion of conformal barycenter \cite{DE}. For instance, one can sample several points from the wrapped Cauchy distribution (\ref{wrapped_Cauchy}). Then, their conformal barycenter is a random point in ${\mathbb B}^2$ with the mean $\alpha = r \Phi$. In fact, this random point is a maximum likelihood estimation of $\alpha$ for the wrapped Cauchy statistical model \cite{CHMR}. This yields conformally natural extensions of wrapped Cauchy densities: if the random points on ${\mathbb S}^1$ are acted on by a M\" obius transformation, their conformal barycenter is transformed by the same transformation. Again, this method can be implemented using simple Kuramoto models \cite{Jacimovic,CEM,CHMR}.

The above argumentation can be extended to probability measures over hyperbolic unit balls ${\mathbb B}^d$.

For yet another alternative, we refer to the paper \cite{NYFK} which presents an inference algorithm over hyperbolic spaces with an application to word embeddings. This algorithm is based on another family, named "hyperbolic normal distributions" in the paper. We find this statistical model pretty inconvenient, mostly because it does not posses any group-invariance properties.

\subsection{Statistical models over orthogonal groups, Grassmannians, homogeneous spaces}

Probability distributions on orthogonal groups or Grassmannians can be obtained from distributions on spheres using various group-theoretic relations and geometric properties.

Statistical models on the group $SO(3)$ can be deduced from models on the sphere ${\mathbb S}^3$ by using the fact that the $SU(2)$ with the group manifold ${\mathbb S}^3$ is the double cover (the $2-1$ group homomorphism) of $SO(3)$. Hence, any family over ${\mathbb S}^3$ can be used for modeling probability distributions over $SO(3)$. 

The idea of generating random orthogonal matrices using von Mises-Fisher distribution on the sphere has been proposed in the old paper \cite{KM}.

For another particular example recall the double cover group homomorphism $SU(2) \times SU(2) \to SO(4)$ which can be used for the construction of families on $SO(4)$ using families on ${\mathbb S}^3$.

In the similar way, one can use homomorphism $G_2^+({\mathbb R}^4) = {\mathbb S}^2 \times {\mathbb S}^2$ for statistical learning on the Grassmannian manifold of 2-dimensional subspaces in the 4-dimensional real space.

We also mention the reference \cite{Chikuse} on statistical ML algorithms on Grassmannians, implemented through the matrix Bingham distribution and matrix Langevin distribution. The later can be obtained from von Mises-Fisher distributions on spheres.

\section{Swarms on manifolds for DL}

We have presented a broad class of models that generate trajectories on various manifolds. Given a particular setup, an appropriate model can be chosen and trained in order to learn an optimal (in a certain sense) trajectory on a specific manifold. Training these dynamical systems boils down to the estimation of parameters: coupling strengths, phase shifts, initial positions of (generalized) oscillators, etc. Noiseless models with global coupling generate trajectories on invariant statistical manifolds (wrapped Cauchy, spherical Cauchy, etc.). On the other hand, some important distributions (von Mises, hyperbolic von Mises, von Mises-Fisher) arise as stationary distributions in Kuramoto models with noise.  

In this Section we briefly discuss ideas on implementation of the corresponding algorithms.

\subsection{Training swarms on manifolds for supervised ML}

Several papers \cite{CK,LP} addressed the problem of reconstructing Kuramoto networks, based on sequential observations of states of oscillators. Algorithms proposed therein can be adapted to ML problems.
An illustrative example is reported in \cite{CK}. Authors consider Kuramoto model (\ref{Kuramoto_123}) with symmetric couplings.
System parameters to be learned are $K_{ij}$ and $\beta_{ij}$.  

Assuming that couplings are symmetric (i.e. $K_{ij}=K_{ji}$ and $\beta_{ij}=\beta_{ji}$) (\ref{Kuramoto_123}) is the Langevin dynamics on $N$-torus for the potential (\ref{energy}).

Substituting $z_j = e^{i \varphi_j}$, the energy function can be written in complex coordinates as
\begin{equation}
\label{energy_complex}
E(z) = z^* {\bf K} z,
\end{equation}
where $z = (z_1,\dots,z_N)$ and ${\bf K}$ is $N \times N$ Hermitian matrix with entries $\kappa_{ij} = K_{ij} e^{i \beta_{ij}}$.

\subsubsection{Maximum likelihood}

The paper \cite{CK} discusses two methods for training (reconstruction) the system (\ref{Kuramoto_123}). The first method is essentially standard ML backpropagation. Given the likelihood function $q(\varphi) = q(\varphi_1,\dots,\varphi_N)$ and the data distribution $p(\varphi) = p(\varphi_1,\dots,\varphi_N)$, the maximum likelihood for the observed data is obtained by differentiating the log-likelihood function and setting the derivative to zero:
\begin{equation}
\label{max_likelihood}
\frac{\partial \langle \log q(\varphi) \rangle_{p(\varphi)}}{\partial \kappa_{ij}} = \left \langle \frac{\partial E}{\partial \kappa_{ij}} \right \rangle_{q(\varphi)} - \left \langle \frac{\partial E}{\partial \kappa_{ij}} \right \rangle_{p(\varphi)}= 0,
\end{equation}
where $\langle \cdot \rangle_{q(\varphi)}$ stands for the mathematical expectation w.r. to distribution $q(\varphi)$. 

Equations (\ref{max_likelihood}) can be solved by the (stochastic) gradient descent method. This algorithm requires approximations of the gradient by iterative sampling from the distribution $q(\varphi)$.

\subsubsection{Score matching}

An alternative method for training Kuramoto networks is score matching \cite{Hyvarinen}. This method has been implemented in \cite{CK} for reconstruction of small networks (consisting of 4-5 oscillators). Introduce the score function:
$$
J_{SM}({\bf K}) = \left \langle \frac{1}{2} |\nabla_\varphi E(\varphi)| |\nabla_\varphi E(\varphi)^T| - \nabla^2_\varphi E(\varphi) \right \rangle
$$ 
where the expectation $\langle \cdots \rangle$ is taken over the data distribution.

Since the energy function (\ref{energy_complex}) is quadratic in data $z$ and linear in system parameters ${\bf K}$, the derivative of $J_{SM}$ is easily calculated. Setting this derivative to zero yields the system of linear algebraic equations for unknowns $\kappa_{ij}$.

\subsubsection{Evolutionary optimization (CMA ES)}

Yet another class of methods for training Kuramoto networks comes from the field of evolutionary optimization and, most notably, the famous CMA ES algorithm \cite{OAAH}. Many experiments demonstrated that CMA ES is competitive with the gradient-based optimization methods when the dimension is not too large (say, the number of parameters is less than one hundred).

Notice, however, that CMA ES performs an update over the Gaussian family and is adapted to Euclidean spaces. This can be suitable for Kuramoto models of the form (\ref{Kuramoto_123}) (the space of $K_{ij}$ is Euclidean, while phase shifts $\beta_{ij}$ are points on the circle). However, when the problem requires learning over the full space of orthogonal matrices, or points on spheres, CMA ES is no longer suitable. In such cases, one can employ evolutionary optimization over manifolds. The stochastic search algorithms over orthogonal groups and spheres can be implemented by performing an update over statistical models considered in the previous Section.

\subsubsection{Reduction the number of parameters by choosing an appropriate model}

In general, Kuramoto networks with $N$ oscillators include approximately $N^2$ parameters (couplings $K_{ij}$ and phase shifts $\beta_{ij}$).
However, depending on the specific task, an appropriate choice of the model greatly reduces the number of parameters. For instance, when learning (coupled) group actions one can exploit symmetries and consider Kuramoto ensembles with several globally coupled sub-ensembles. This reduces the number of parameters to the order of $k^2$, where $k$ is a number of sub-ensembles. In the case of probabilistic modeling, this boils down to learning mixtures of $k$ wrapped Cauchy (or spherical Cauchy) distributions.

For Lorentz groups, parameters to be learned are global couplings and initial positions of oscillators. For instance, suppose that our goal is to learn $k$ coupled actions of the M\" obius group $PSU(1,1) \simeq SO(2,1)$. For this goal, one can set the Kuramoto model (\ref{Kuramoto_sub-ensemble}) with $k$ sub-ensembles, and learn 5 parameters for each sub-ensemble, namely: a) internal coupling strength $K_{ii}$; b) internal phase shift $\beta_{ii}$ and; c) initial positions for three oscillators, since the M\" obius transformation is uniquely determined by its action on three points on $S^1$. In addition, the algorithm needs to learn $k(k-1)/2$ couplings between sub-ensembles. In total, this makes $k(k+9)/2$ parameters for learning $k$ coupled trajectories in M\" obius groups. This will be explained in more detail on the illustrative example below.

\subsection{Swarms on manifolds and directional statistics in RL}

References on RL on non-Euclidean spaces are very sparse. Researchers only recently started to apply RL techniques to the learning over data sets with non-Euclidean geometries. Reward-based update of deterministic and stochastic policies on Riemannian manifolds has potential applications in different fields. Hence, Markov decision processes with non-Euclidean space of states (and actions) present an important challenge for ML. However, rather than developing the general theory, it seems more advantageous to start with some paradigmatic examples in order to elaborate conceptual approaches. Two recent studies \cite{JA,ASSAKA-D} considered RL algorithms on spheres and special orthogonal groups. Both studies proposed the Bingham distribution for implementation of stochastic policies. 

Parametrization of stochastic policies by spherical Cauchy and von Mises-Fisher distributions in RL is still to be explored. In our point of view, the idea of using spherical Cauchy distributions in RL has a great potential, given all nice properties of this family. We will discuss this idea on some illustrative examples in the next Section.

We also mention a very recent paper \cite{CCBH} on hyperbolic RL. This paper proposes deep RL algorithms for designing more efficient policies that model latent representations in the hyperbolic space. Like most of ML algorithms in hyperbolic geometries this method exploits the framework of gyrovector spaces for operations in hyperbolic spaces.

\subsection{Swarms on manifolds and directional statistics for unsupervised ML}

There are several influential studies \cite{BDGS,GY} on clustering spherical data using mixtures of von Mises-Fishers. Banerjee et al. \cite{BDGS} introduce two expectation maximization algorithms for estimating the mean and concentration parameters for such mixtures and propose spherical $k$-means clustering algorithm.

The paper \cite{CJ-NC} reports experiments with real Kuramoto models on spheres for the simultaneous clustering of Euclidean data (encoded in frequency matrices) and hierarchical data (encoded in the coupling network). 

\subsection{Statistical models for the latent space}

Variational autoencoders (VAE) are among the most prominent classes of probabilistic graphical models in DL. VAE consists of two NN's and the latent space. The first network (encoder) embeds input vectors from a high-dimensional vector space $X$ into a lower-dimensional latent space $Z$; the second network (decoder) maps $Z$ into a vector space $Y$ of the output data. Encoder and decoder networks are trained simultaneously, with the objective of maximizing the likelihood of the observed data $x \in X$ by learning over a parametrized probability distributions $p(x | \theta)$ in $Z$. In brief, architecture of VAE can be illustrated by the following simple diagram: 
$$
X \xmapsto{\phi_\theta} Z \xmapsto{\psi_\theta} Y.
$$
Here, $\phi_\theta$ and $\psi_\theta$ denote the maps that are usually implemented through NN's.

The encoder network learns probabilistic representations of input vectors from $X$. Once the training is completed, this network becomes unnecessary, and the new data is generated by sampling from a chosen probability distribution $p(x|\theta) \in Z$ and mapping the samples to $Y$ via the decoder $\psi_\theta$.
 
An essential issue in designing VAE is the structure of the latent space. In principle, any family of probability distributions (statistical manifold) can be chosen to model the input data. However, statistical manifolds are not equally convenient for learning algorithms. First, VAE generates the new data through sampling from $p(x | \theta)$. This imposes an obstacle for the training, because sampling is not a differentiable operation and one can not backpropagate through this operation. Second, VAE is trained in order to minimize the reconstruction error between the output and input data. This objective is achieved by simultaneous maximization of the log-likelihood of the observed data and minimization of the K-L divergence between the approximate posterior $p_\theta(\cdot|x)$ and the exact posterior $q_\theta(\cdot|x)$. This is achieved through the maximization of the evidence lower bound (ELBO). Hence, reasonably simple expression for the K-L divergence is essential for the efficient training. 

The standard choice proposed in the first versions of VAE \cite{KW,KRMW} is the family of Gaussian distributions ${\cal N}(a,\Sigma)$. In order to avoid costly calculations with the full covariance matrix, the model is further simplified by taking Gaussian family ${\cal N}(a,\sigma^2 I)$ with a diagonal covariance matrix. 

The Gaussian family has several advantages. The most important is the group-invariance property (P4). This property allows to circumvent the obstacle of differentiating through the sampling operation using the so-called {\it reparametrization trick} (sometimes also referred to as {\it stochastic backpropagation}) \cite{KW}. This trick exploits the fact that $y \sim N(a,\Sigma)$ can be sampled by sampling a vector $x \sim N(0,I)$ and transforming it as $y = \Sigma^{1/2} x + a$. Hence, one only needs samples from the prior ${\cal N}(0,I)$ in order to calculate gradients w.r. to $a$ and $\Sigma$.

The second advantage of the Gaussian family is property (P2). A simple expression for the K-L divergence facilitates maximization of ELBO. 

In whole, VAE's with the Gaussian latent space (named normal, or Gaussian, VAE and denoted by ${\cal N}$-VAE) are standard and the most popular model. Following the seminal papers \cite{KW,KRMW} on VAE several alternative versions of VAE have been proposed and attracted a lot of interest. One alternative are VAE's with the latent space consisting of the family of categorical distributions (i.e. distributions over a finite set of outcomes). It has been argued that such an architecture named categorical VAE (we use an abbreviation ${\cal CAT}$-VAE) can be advantageous in many setups. The problem of backpropagation through the sampling is circumvented using the "Gumbel-softmax" trick \cite{JGP}. This trick is based on the fact that categorical distributions can be approximated with the family of absolutely-continuous Gumbel distributions by gradually annealing the "temperature" parameter. Hence, training of ${\cal CAT}$-VAE is based on the backpropagation through the family of Gumbel distributions and annealing the temperature in order to approximate the targeted categorical distribution. Maximization of ELBO in ${\cal CAT}$-VAE is not difficult, since there is a simple explicit expression for the K-L divergence between two categorical distributions.

Further experiments explored VAE architectures with von Mises-Fisher distributions in the latent space \cite{XD,DFdCKT}. These architectures have been named spherical VAE and denoted ${\cal S}$-VAE. We prefer the abbreviation ${\cal VMF}$-VAE in order to emphasize that they assume the von Mises-Fisher family. A difficulty in training ${\cal VMF}$-VAE is the backpropagation through the sampling. The reparametrization trick is realized through the rejection-acceptance sampling \cite{NRLB}. It has been argued that ${\cal VMF}$-VAE is able to capture directional data \cite{DFdCKT}, or word embeddings \cite{XD}, where ${\cal N}$-VAE fails.

Underline that all three families proposed so far contain natural priors. In ${\cal N}$-VAE, the standard Gaussian distribution ${\cal N}(0,I)$ serves as a prior. In ${\cal CAT}$-VAE the prior is a uniform distribution over a finite set (that is: $p_1 = \cdots = p_N = 1/N$ for $N$ possible outcomes). Finally, the von Mises-Fisher family contains the uniform distribution on ${\mathbb S}^{d-1}$ for zero value of the concentration parameter. 

We point out that families of wrapped Cauchy and spherical Cauchy provide more convenient options for the latent space in VAE (although we are not able to present experiments at the moment). In ${\cal WC}$-VAE the latent space would be equipped with wrapped Cauchy distributions on the torus ${\mathbb T}^d$. This family is parametrized by $d$ points inside the hyperbolic disc. In ${\cal SC}$-VAE, the latent space consists of spherical Cauchy distributions on ${\mathbb S}^{d-1}$. Parameter of this family is a single point inside the hyperbolic ball.

The main advantage of ${\cal WC}$-VAE and ${\cal SC}$-VAE is the group-invariance property (P4), allowing for an easy implementation of reparametrization trick. One can sample from the uniform distribution on ${\mathbb S}^{d-1}$ and map the sample using hyperbolic isometries of the unit ball. In such a way we obtain samples from any spherical Cauchy distribution. Hyperbolic isometries $g(w)$ are invertible and differentiable (w.r. to $w$) transformations. Differentiating with respect to $w$ boils down to computing generators of the group of hyperbolic isometries \cite{LMS}.

Recently, new VAE architectures with latent dynamics described by continuous-time dynamical systems have been investigated on sequential data (irregular time series) \cite{RCD,YHL}. This idea can be naturally extended to VAE's with spherical (or toroidal) latent space. Learning of the (sequential) data representations in the latent space can naturally be implemented through the training of the Kuramoto networks. As explained above, in such a way we can implement learning over families $vMF, sphC$ and $wC$. In conclusion, VAE architectures with wrapped and spherical Cauchy distributions (along with learning embeddings into the latent space by training Kuramoto models) seem like a very promising idea, but is still to be implemented and tested on real-life problems.

\subsection{Kuramoto models for learning (coupled) actions of Lie groups}

One of the main points of the present study is that many important ML problems are naturally stated as learning coupled actions of certain Lie groups. These include special orthogonal and unitary groups, as well as actions of the Lorentz groups for learning in hyperbolic geometries. In the present subsection we briefly discuss problems of this kind and conceptual approaches to them.

\subsubsection{Non-Abelian Kuramoto models for learning (coupled) actions of orthogonal and unitary groups}

Problems of learning optimal (coupled) rotations are common in robotics, when modeling motions. Apparent examples of this kind are rotations of linked robot's arm with $k$ joints. If the arm moves in the plane, this yields $k$ coupled actions of $SO(2)$. In the three-dimensional space, these are coupled actions of $SO(3)$. 

The problem of learning actions of $SO(d)$ for $d>3$ and $SU(n)$ for $n>2$ is relevant in computational physics and some other fields.

The system (\ref{Kuramoto_SO(n)}) with $k$ oscillators is a convenient model of $k$ coupled rotations in the $n$-dimensional vector space. Parameters to be learned are coupling strengths $K_{ij}$.
This model depends on a small number of parameters and can be trained efficiently. The obvious drawback is the limited representative power. There are several ways of increasing representative power, but they inevitably entail increased number of parameters. For instance, one can include several oscillators for each rotation.

Alternative models for learning $SO(3)$ rotations are provided by models (\ref{Kuramoto_sphere_real}) and (\ref{Kuramoto_sphere_complex}) on ${\mathbb S}^3$.

\subsubsection{Kuramoto models on spheres for learning (coupled) actions of Lorentz groups}

Computational physics is one of many fields that are being revolutionized with the advent of DL. Since Minkowski spacetime and its symmetries are one of central concepts in mathematical physics, DL models involving Lorentz groups have been investigated within this field \cite{BAORMK}.  

However, the significance of ML on Lorentz groups is not restricted to physical applications. With rapid developments in Geometric DL, the importance of hyperbolic geometries in a very wide range of setups is widely recognized. 
Many data sets are naturally represented by points (or probability distributions) over hyperbolic spaces, such as unit balls ${\mathbb B}^d$ or their products ${\mathbb B}^d \times \cdots \times {\mathbb B}^d$. Learning over hyperbolic data can be realized through learning coupled actions of isometries in the corresponding spaces. Recall that groups of hyperbolic isometries of unit balls are isomorphic to Lorentz groups $SO(d,1)$. This motivated recent experiments with Lorentz groups for deep learning of sequential hierarchical data \cite{NK2,LLSZ}. Therefore, Lorentz groups are potentially relevant for essentially all ML algorithms dealing with hierarchical data. 

Kuramoto models with several sub-ensembles (\ref{Kuramoto_sub-ensemble}) and (\ref{Kuramoto_sub_ensemble_sphere}) provide a natural framework for encoding (sequential) coupled actions of hyperbolic isometries in the unit disc and unit balls, respectively. Hence, these models can be used for learning data in hyperbolic spaces. Having in mind omnipresence of hyperbolic data, we believe that this is potentially the most promising field of applications of Kuramoto models in DL.

In order to clarify this general idea, we discuss a particular example. Suppose that we deal with data represented by points on the $d$-dimensional torus ${\mathbb T}^d = {\mathbb S}^1 \times \cdots \times {\mathbb S}^1$ (or in its interior ${\mathbb B}^2 \times \cdots \times {\mathbb B}^2$). Assume that the objective is to learn $d$ isometries of ${\mathbb S}^1$ (or ${\mathbb B}^2$) that transform points into an optimal (according to a criterion, which may be known or black-box) configuration. In addition, these isometries depend on each other. This can be treated as an optimization problem over the product ${\mathbb G} \times \cdots \times {\mathbb G}$ of $d$ M\" obius groups. The system (\ref{Kuramoto_sub-ensemble}) with $d$ sub-ensembles provides an appropriate model for encoding the data of this kind. 

In theory, this is the learning problem over orbits of the $3d$-dimensional Lie group ${\mathbb G} \times \cdots \times {\mathbb G}$, hence learning $3d$ parameters could be sufficient. These parameters are three initial phases $\varphi_i^k(0)$, where $i=1,2,3$ and $k=1,\dots,d$ for each sub-ensemble. 

In practice, it might be reasonable to learn couplings $K_{ij} = K_{ji}$ and phase shifts $\beta_{ij} = \beta_{ji}$ where $i,j = 1,\dots,d$. This makes $d(d+1)$ parameters. Together with $3d$ initial phases, this sums up to $d(d+4)$ parameters. Emphasize that this is a reasonably low number, taking into account the complexity of the problem. Moreover, the whole approach naturally applies to problems with sequential data.

The idea can be straightforwardly extended to learning on products of spheres or balls, by training the model (\ref{Kuramoto_sub_ensemble_sphere}). Underline that training of (\ref{Kuramoto_sub_ensemble_sphere}) must be based on optimization algorithms over Riemannian manifolds, because only coupling strengths are Euclidean. The remaining parameters belong to special orthogonal groups (phase shifts) and to spheres (initial points).

In general, models (\ref{Kuramoto_sub-ensemble}) and (\ref{Kuramoto_sub_ensemble_sphere}) are significant for learning in hyperbolic geometries. An advantage of this approach is that it avoids mathematical apparatus (involving exponential map, gyrovector spaces and M\" obius addition) that is frequently used in ML algorithms over hyperbolic data \cite{TBG,GBH,NK,NK2}. 

The idea explained here raises important issues regarding representative power of the proposed models (\ref{Kuramoto_sub_ensemble_sphere}). This can be mathematically stated as the controllability  problem over the Lie groups $SO(d,1) \times \cdots \times SO(d,1)$, where couplings, phase shifts and initial positions are regarded as controls. Namely, the question is if we can obtain each Lorentz transformation by varying coupling strength, the phase shift and three initial points. 
This question is quite involved and is still to be investigated using the general control theory on Lie groups \cite{Sachkov}. Local controllability results seem to follow from the general theory. However, results on global controllability demand a careful and demanding analysis.   

\subsection{Grassmannian shallow and deep learning}

Some data sets are naturally embedded into Grassmannian manifolds \cite{HSSL}. This fact motivated research efforts in optimization and learning over Grassmannian manifolds. Applications studied so far are mainly in image classification and recognition. 

We refer to \cite{ZZHH} for some applications of shallow and deep Grassmannian learning. The problems of shallow learning are typically stated as optimization problems over Grassmannian manifolds. These include classification, spectral clustering of high-dimensional data, and low rank matrix completion. The DL problems solved by Grassmannian methods include transfer learning (generalizing the knowledge from one field to another) and feature extraction. Applications are in Image-set/Video based recognition and classification, MIMO communication, recommender systems \cite{ZZHH}.

The paper \cite{HWG} proposes deep NN architecture working with the Grassmannian input data. Weights of such NN are matrices and the training is implemented by backpropagation over the matrix groups. 

Systems of Riccati ODE's on Grassmannians \cite{Zelikin,Sarlette} (see Section 4.3) can be used for ML on these manifolds. Such an approach has been reported in \cite{AMS1} with applications to some problems of shallow Grassmannian learning.

\subsection{Ensembles of coupled oscillators in ML: Beyond Kuramoto models}

For the sake of completeness of the exposition, we mention some previous ML experiments with ensembles of coupled phase oscillators. Underline that most of these studies experimented with dynamical systems which do not belong to Kuramoto models. 

\subsubsection{Ensembles of coupled phase oscillators as associative memories}

 Several studies proposed to regard ensembles of coupled oscillators as associative memories \cite{HI1,HI2,NLH}. This proposal was inspired by the Hopfield model, since ensembles of phase oscillators can be treated as generalizations of the famous Ising model. Ensembles of phase oscillators with symmetric couplings yield gradient flows on tori.

Models proposed as associative memories {\bf are not} Kuramoto models, since they include couplings through higher-order harmonics. Hence, they are not systems of geometric matrix Riccati ODE's.
 
One could think about designing associative memories based on extensions of ensembles of coupled oscillators to higher-dimensional manifolds. This might sound as a promising idea, but the corresponding theory is non-existent at the moment. Indeed, couplings through higher harmonics have been studied only for the ensembles of the classical phase oscillators, but not for generalized oscillators on spheres and other manifolds. Therefore, further investigations in this direction does not seem very promising, due to the total lack of theoretical background.     

\subsubsection{Swarms with plastic synapses for ML}

An intriguing field of study in ML are NN architectures with plastic synapses (adaptive weights). The whole idea is, to a great extent, inspired by Neuroscience. We refer to the recent papers \cite{MCS,SSR} for fairly comprehensive reviews of NN's with plastic synapses. Networks of this kind can be trained through backpropagation \cite{MCS}. 

The central paradigm in networks with plastic synapses is the Hebbian learning rule. This rule, often summarized as "cells that fire together, wire together", imposes that synapses between two neurons get stronger, as their states get closer. 

Adaptive synapses obeying Hebbian rules are easily and naturally implemented in networks of Kuramoto oscillators \cite{HNP}. Furthermore, it is straightforward to extend this model by introducing Kuramoto models on spheres with adaptive couplings \cite{CJ-SCL}. Many models of this kind exhibit potential dynamics.

One can enrich Kuramoto models on orthogonal groups and spheres by assuming that the couplings are adaptive and setting up the learning rule. In order to increase the representative power one can combine Hebbian and anti-Hebbian learning rules. In such a way one can construct the models representing coupled actions of the group $SO(d)$ or complicated probability distributions on spheres. 

Notice that systems with adaptive couplings are not suitable for encoding hyperbolic isometries. Hence, actions of the Lorentz groups can not be modeled in this way. 

\section{Examples}

We present several illustrative problems in order to support the main points of this article. We intentionally choose relatively simple examples in order to illustrate how particular problems can be addressed by appropriately chosen models from the proposed general framework.   

For the sake of brevity of exposition we omit technical details on implementation. The details and results will be presented elsewhere.

\subsection{Wahba's problem}

Wahba's problem \cite{Wahba} can be regarded as a classical ML problem over Riemannian manifolds, although it has been stated in 1965 before the advent of ML. 
The problem consists in finding a rotation in the three-dimensional space ($SO(3)$ matrix) from a set of noisy (possibly weighted) observations. It can be stated as follows.

Given $m>2$ input vectors $v_1,\dots,v_k$ in ${\mathbb R}^3$ and $m$ noisy observations $w_1,\dots,w_k$ find a rotation matrix $R$ that minimizes the total disagreement between vectors $R v_i$ and $w_i$. The process can be written in the following form
$$
w_i = R v_i + \xi_i, \quad i=1,\dots,m
$$
where $\xi_i$ are mutually independent realizations of white noise.

The disagreement function is usually assumed to be of the following form:
$$
J(R) = \frac{1}{2} \sum_{i=1}^m a_i || w_i - R v_i||^2, \mbox{  where  } a_i > 0 \mbox{  are the weights.}
$$

Hence, this is an optimization problem over $SO(3)$. It might be considered as a problem of "shallow" ML. There are efficient solutions which exploit the apparatus of linear algebra (for instance, SVD matrix decomposition). 

Recent papers \cite{JA,ASSAKA-D} on RL on manifolds tackled Wahba's problem in order to illustrate stochastic policies over $SO(3)$. Both papers experimented with the Bingham policy parametrization. 

In contrast, we implemented the spherical Cauchy policy parametrization over the sphere ${\mathbb S}^3$ and exploited the double cover mapping from ${\mathbb S}^3$ to $SO(3)$. This is a convenient choice, since it requires only a single four-dimensional parameter of the spherical Cauchy family to be learned. This parameter is a point $\zeta = \rho \cdot \mu$ in the hyperbolic ball ${\mathbb B}^4$ (see (\ref{spherical_Cauchy})). In this way the optimization problem on $SO(3)$ is replaced by the problem over the hyperbolic ball ${\mathbb B}^4$. Notice that the choice of the spherical Cauchy family entails the necessity of taking into account the problem of "gimbal loop". However, it is not difficult to circumvent this potential error, for instance, by mapping all the points to the upper hemi-sphere.

We have also experimented with implementations of the spherical Cauchy policies via the Kuramoto model on ${\mathbb S}^3$ with global coupling. In such a way, the problem is turned into optimization of parameters $\omega, K$ and $\beta$ of the Kuramoto model over the space ${\mathbb R}_+ \times {\mathbb R} \times {\mathbb S}^1$.

Alternatively, Wahba's problem can be addressed by the non-Abelian Kuramoto model (\ref{Kuramoto_SO(n)}) on $SO(3)$ with the global coupling $K_{ij} \equiv K$.

The remaining two families exposed in subsection 5.2 (von Mises-Fisher and Bergman-Cauchy) can also be used for the policy parametrization over ${\mathbb S}^3$. We believe that Cauchy and Bergman-Cauchy are more convenient options than the von Mises-Fisher and Bingham because of their group-invariance properties.


\subsection{Linked robot's arm (planar rotations)}

Suppose that we have a set of observations of the linked robot's arm with $m-1$ joints. Mathematically, this is formalized as a problem of learning $m$ coupled actions of the group $SO(2)$. Recent study \cite{RPRAKSC} experimented with three types of normalizing flows on the torus with applications to this problem.

\subsubsection{Deterministic policy for learning coupled planar rotations}
 
The deterministic policy can be stated as an optimization problem over the torus ${\mathbb T}^m = {\mathbb S}^1 \times \cdots \times {\mathbb S}^1$.
 
Coupled planar rotations can be encoded into the dynamics (\ref{Kuramoto_123}) with $m$ oscillators. The system (\ref{Kuramoto_123}) contains $m(m-1)$ parameters $K_{ij}$ and $\beta_{ij}$ to be learned. If the number of joints is reasonably small (say, less than ten), evolutionary optimization algorithms (such as CMA ES) perform pretty well.

\subsubsection{Stochastic policy for learning coupled planar rotations}

Probabilistic modeling of coupled rotations is more involved. Appropriate statistical models for this data set are provided by probability distributions on the torus ${\mathbb T}^m$ discussed in subsection 5.1. The simplest option is to choose the family of probability distributions on torus whose marginals on circles are wrapped Cauchy. Such a policy can be implemented through training of the Kuramoto model (\ref{Kuramoto_sub-ensemble}) with $m$ sub-ensembles. Initial positions of oscillators are sampled from the uniform distribution on ${\mathbb T}^m$, and the model is trained in order to find the probability distribution from this family which approximates observations. Since each wrapped Cauchy distribution is parametrized by a point in the hyperbolic disc ${\mathbb B}^2$, implementation of this stochastic policy can be formalized as an optimization problem over the product of hyperbolic discs ${\mathbb B}^2 \times \cdots \times {\mathbb B}^2$.

The simplicity of such stochastic policy implies limited representative power. Indeed, this model can learn only distributions whose marginals are symmetric and unimodal. This can be sufficient if the motions are not very complicated.

In order to increase the representative power, one can divide each sub-ensemble into several sub-sub-ensembles thus increasing the number of parameters. In such a way, the system can learn distributions on ${\mathbb T}^m$ whose marginals are mixtures of wrapped Cauchy.

Another option is to choose the Kato-Jones family (\ref{Kato-Jones}). More precisely, one would learn a probability distribution on ${\mathbb T}^m$ whose marginals are Kato-Jones. This distribution can be approximated by the two-stage dynamics, with the first stage transforming marginal uniform into the von Mises distributions, and the second stage transforming von Mises into Kato-Jones. We point out the recent study \cite{NKNJ} which proposes mixtures of two Kato-Jones for the density estimation on the circle. 

This approach can naturally be applied to the sequential (temporal) data. Therefore, the Kuramoto model can be trained to predict and imitate motions of the linked robot's arm in real time.

\subsection{Linked robot's arm (spatial rotations)}

When dealing with coupled rotations in the three-dimensional space, one can use non-Abelian Kuramoto model (\ref{Kuramoto_SO(n)}) on $SO(3)$ or one of the two Kuramoto models on the three sphere (\ref{Kuramoto_sphere_real}) or (\ref{Kuramoto_sphere_complex}).

For probabilistic modeling, the most convenient choice seems to be the family of probability distributions on the $m$-product of spheres ${\mathbb S}^3 \times \cdots \times {\mathbb S}^3$, where the marginals are spherical Cauchy distributions on ${\mathbb S}^3$. This family is approximated by the Kuramoto model (\ref{Kuramoto_sub_ensemble_sphere}) with $m$ sub-ensembles (with the initial positions sampled from the uniform distribution on ${\mathbb S}^3 \times \cdots \times {\mathbb S}^3$). 

Again, such a choice is convenient, but implies a limited representative power with unimodal and symmetric marginal distributions on ${\mathbb S}^3$. In order to increase the representative power, one can divide each sub-ensemble into several sub-sub-ensembles thus capturing distributions on ${\mathbb S}^3 \times \cdots \times {\mathbb S}^3$ whose marginals on ${\mathbb S}^3$ are multi-modal.

Alternatively, one might opt to use Bingham family (\ref{Bingham}). An advantage of this family is that it is antipodally symmetric, thus avoiding the gimbal loop problem. However, such a choice has many drawbacks, since Bingham family does not posses group-invariance properties and can not be generated by any swarm or Kuramoto model. Moreover, there is too many parameters to learn. In our point of view the ratio between simplicity and the representative power is not favorable for the Bingham parametrization policy.

\subsection{Embedding multilayer complex networks (Learning coupled actions of Lorentz groups)}

Random geometric graph are constructed by randomly placing $N$ nodes in some metric space and assuming that the probability that two nodes are linked decreases with the distance between them \cite{Penrose}. This point of view on graphs has many advantages. For instance, one can treat graphs as realizations of a certain statistical law. In particular, two different graphs may be particular realizations of the same statistical law.

Some topologies, such as Erd\" os-R\' enyi model \cite{ER}, are obtained by sampling the nodes from the uniform distribution in a compact subset of the Euclidean space \cite{Penrose}.  

However, complex networks appearing in Science, Engineering and everyday life are not completely random. The characteristic property of such networks is that degrees of nodes are distributed according to the power law \cite{Newman,KPKVB}. This property implies that such networks are naturally embedded into hyperbolic, rather than Euclidean space \cite{KPKVB}.

Hence, complex networks belong to a very broad class of data sets that are represented by point clouds and probability distributions in hyperbolic spaces. Notice that the same applies to essentially all data sets with hierarchical structure. However, many real-life data sets encompass multiple types of relationships, which are represented by multilayer (multidimensional, multiplex) networks \cite{KABGMP}. For example, in natural languages words may be related at several different levels. Multilayer complex networks are naturally represented in the product of hyperbolic spaces.

In order to faithfully embed large networks, two-dimensional disc ${\mathbb B}^2$ may be insufficient, because the distances between nodes can become very small. In such cases one can represent the data in higher-dimensional hyperbolic balls \cite{Munzner}. Each level of the network is represented by the data cloud in the hyperbolic ball ${\mathbb B}^d$. By acting on the ball with hyperbolic isometries we obtain the same geometric network. Algorithms of embedding complex networks in hyperbolic spaces are well studied and work efficiently if the network is not very big \cite{BFKL,MTCBC,G-PASB}.   

Given a network with $m$ layers, there are efficient methods of embedding layers into balls ${\mathbb B}^d$. Group of hyperbolic isometries acting on ${\mathbb B}^d$ does not affect the layer. However, relationships between nodes belonging to different layers are not taken into account. In order to fix that we need to learn coupled actions of $m$ hyperbolic isometries, each of them acting on one ball. Introducing the loss function, we can use the model (\ref{Kuramoto_sub_ensemble_sphere}) in order to learn the optimal trajectory in the $m$-product of hyperbolic isometries ${\mathbb Q} \times \cdots \times {\mathbb Q}$. We can summarize the method as following:

1. Embed each of $m$ layers into hyperbolic balls ${\mathbb B}^d$.  

2. Introduce the loss function between layers (point clouds in balls). This function represents relationships between nodes belonging to different layers. Suppose that each point cloud consists of $p$ points and there are $r$ layers. Then the loss function is a sum of $p^2 r(r-1)/2$ terms, each of them equal to zero or one. If two nodes belonging to different layers are not connected, then the hyperbolic distance between the corresponding points in ${\mathbb B}^d$ should be larger than a certain threshold value. 

3. Sample $d(d+1)/2$ points from the uniform distribution on each of $m$ spheres ${\mathbb S}^{d-1}$.

4. Simulate dynamics (\ref{Kuramoto_sub_ensemble_sphere}) with $m$ sub-ensembles and initial positions of oscillators sampled in the previous step.

5. Train the model (\ref{Kuramoto_sub_ensemble_sphere}) by learning initial positions of oscillators, coupling strengths $K_{kl}$ and phase shifts $Q_{il}$ in order to minimize the loss function.

6. Once the training is completed, M\" obius transformations acting on each disc are learned. Extend these transformations from the circle to the disc. Map the points via learned transformations in each disc.

\section{Conclusion}

We presented a broad framework for ML on non-Euclidean spaces based on swarms on manifolds. By swarms we mean a class of dynamical systems that can be regarded as higher-dimensional extensions of the classical Kuramoto model. The proposed approach is based on solid theoretical foundations, established in the previous 25 years, mostly within physics of complex systems and geometric consensus theory. We leverage theoretical knowledge to choose appropriate models for specific problems, thus greatly decreasing the dimensionality and increasing efficiency and transparency of algorithms.   

The most significant property of swarming dynamics on manifolds is the ability to encode actions of transformation groups. Most notably, they are capable of learning coupled actions of rotation groups, Lorentz groups, and many others. This allows for an efficient learning and prediction of spherical and hyperbolic data.

Relations with directional statistics are essential for probabilistic modeling and statistical ML algorithms. Probabilistic ML on non-Euclidean data is at a very early stage, but significance and wide range of applications of this field is now widely recognized. In Section 5 we have presented an overview of families of probability distributions that can serve as statistical models in problems of this kind. We have also discussed advantages and drawbacks of each family. Some of these families (such as wrapped and spherical Cauchy) have not been used in ML so far, despite their convenience and favorable properties. We exposed the relation between some of these families with various Kuramoto models. In general, we argue that families that can be approximated (i.e. sampled from) using Kuramoto models are typically the most convenient statistical models in most setups.

In Section 6 we discussed approaches to the training of such networks. Since Kuramoto models have not been applied in ML so far, methods of their training are not sufficiently investigated. However, standard optimization methods used in ML can be easily adapted to training of Kuramoto networks. These include (stochastic) gradient descent, score matching and evolutionary optimization. 

Although theoretical framework for ML via using swarms on manifolds is mainly completed, there are still some remaining caveats. One important issue is related to the representative power in learning actions (generators) of the Lorentz groups. Mathematically, this question can be formalized as a controllability problem on Lie groups \cite{Sachkov}.

In Section 7 we discussed several illustrative problems of geometric ML that can be treated by the approach proposed here. We explained methods for dealing with such problems, but did not provide any simulation results. This is omitted for the sake of brevity, but also because we have only started with experiments. It requires significant efforts to design the loss functions, implement the training of various Kuramoto networks and compare with other approaches. Nevertheless, we are convinced that the idea merits the efforts, since many important problems can be dealt with using this universal and flexible framework.






\end{document}